\documentclass{article}

\usepackage[final]{neurips_2025}

\usepackage[utf8]{inputenc} 
\usepackage[T1]{fontenc}    
\usepackage{hyperref}       
\usepackage{url}            
\usepackage{booktabs}       
\usepackage{amsfonts}       
\usepackage{nicefrac}       
\usepackage{microtype}      
\usepackage{xcolor}         
\usepackage{amssymb}
\usepackage{amsmath}

\usepackage{algorithm}
\usepackage{algpseudocode}
\algrenewcommand\algorithmicindent{.9em}%
\usepackage{array}
\usepackage[caption=false,font=normalsize,labelfont=sf,textfont=sf]{subfig}
\usepackage{textcomp,caption}
\usepackage{stfloats}
\usepackage{verbatim}
\usepackage{graphicx}
\usepackage{tabularx}
\usepackage{color}
\usepackage{multirow}
\usepackage{colortbl}
\usepackage{orcidlink} 
\usepackage{soul}
\usepackage{ulem}
\newcommand{\vcenteredbox}[1]{\begingroup
\setbox0=\hbox{#1}\parbox{\wd0}{\box0}\endgroup}

\title{Learning Temporal 3D Semantic Scene Completion \\via Optical Flow Guidance}

\author{
	Meng Wang$^{1}$$^{*}$ \hspace{1em}
	Fan Wu$^{1}$\thanks{Equal contributed. $\dagger$Corresponding authors.} \hspace{1em}
	Ruihui Li$^{1}$$^{\dagger}$ \hspace{1em}
	Yunchuan Qin$^{1}$ \hspace{1em}
	Zhuo Tang$^{1}$$^{\dagger}$ \hspace{1em}
	Kenli Li$^{2,1}$
	\\
	\footnotesize$^1$College of Computer Science and Electronic Engineering, Hunan University\\
	\footnotesize$^2$State Key Laboratory of Advanced Design and Manufacturing Technology for Vehicle, Hunan University\\
  \footnotesize\texttt{\{willem, wufan, liruihui, qinyunchuan, ztang, lkl\}@hnu.edu.cn} \\
	\footnotesize Project Page: \href{https://github.com/willemeng/FlowScene}{https://github.com/willemeng/FlowScene}
}

\begin{document}

\maketitle

\begin{figure}[ht]
    \begin{center}
    \vspace{-3em}
\centerline{\includegraphics[width=1\linewidth]{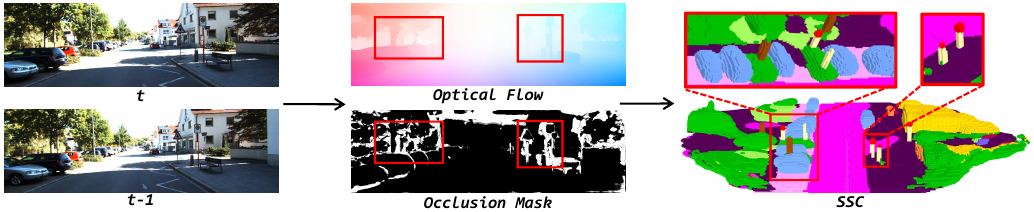}}
        \vspace{-0.3em}
        \caption{Given the temporal RGB images as input, our method can perform temporal modeling based on the corresponding optical flow and occlusion mask, and predict semantic scene completion for all voxels in 3D space.}
        \label{fig:figure1}
    \end{center}
    \vspace{-2.2em}
\end{figure}

\begin{abstract}
3D Semantic Scene Completion (SSC) provides comprehensive scene geometry and semantics for autonomous driving perception, which is crucial for enabling accurate and reliable decision-making. However, existing SSC methods are limited to capturing sparse information from the current frame or naively stacking multi-frame temporal features, thereby failing to acquire effective scene context. These approaches ignore critical motion dynamics and struggle to achieve temporal consistency. To address the above challenges, we propose a novel temporal SSC method FlowScene: Learning Temporal 3D Semantic Scene Completion via Optical Flow Guidance. By leveraging optical flow, FlowScene can integrate motion, different viewpoints, occlusions, and other contextual cues, thereby significantly improving the accuracy of 3D scene completion. Specifically, our framework introduces two key components: (1) a Flow-Guided Temporal Aggregation module that aligns and aggregates temporal features using optical flow, capturing motion-aware context and deformable structures; and (2) an Occlusion-Guided Voxel Refinement module that injects occlusion masks and temporally aggregated features into 3D voxel space, adaptively refining voxel representations for explicit geometric modeling.
Experimental results demonstrate that FlowScene achieves state-of-the-art performance, with mIoU of 17.70 and 20.81 on the SemanticKITTI and SSCBench-KITTI-360 benchmarks. 
\end{abstract}
\section{Introduction}
 \label{sec:intro}
One of the key challenges in autonomous driving is 3D scene understanding, which involves interpreting the spatial layout and semantic properties of objects within the scene. The ability to perceive and accurately interpret 3D scenes is essential for making safe and informed driving decisions. Recently, the 3D Semantic Scene Completion (SSC) task~\cite{song2017semantic,roldao2020lmscnet} has gained significant attention in autonomous driving, as it enables the joint inference of geometry and semantics from incomplete observations.

Most existing SSC methods~\cite{rist2021semantic,zhang2018efficient,guo2018view,li2020aicnet,roldao2020lmscnet,yan2021sparse} rely on input RGB images along with corresponding 3D data to predict volume occupancy and assign semantic labels. However, the dependence on 3D data often requires specialized and costly depth sensors, which can limit the broader applicability of SSC algorithms. Recently, many researchers~\cite{cao2022monoscene,zhang2023occformer,li2023stereoscene,huang2023tri} have investigated camera-based approaches to reconstruct dense 3D geometric structures and recover semantic information, offering a more accessible alternative.

\begin{figure}[t]
\begin{center}
   \includegraphics[width=0.65\linewidth]{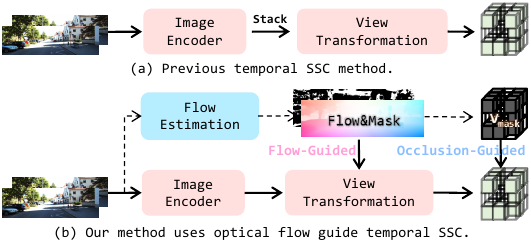}
\end{center}
   \caption{Our method uses optical flow guide temporal SSC versus the previous temporal SSC method.}
\label{fig:figure2}
\vspace{-5mm}
\end{figure}

Previous camera-based SSC methods~\cite{jiang2024symphonize,CGFormer,li2023stereoscene} typically rely on the limited observations available in the current frame to recover 3D geometry and semantics. Later, some researchers~\cite{li2023voxformer,li2024htcl,mei2024sgn,wang2024HASSC} stacked historical temporal features or aligned features with estimated camera poses to enrich contextual information, as shown in Figure~\ref{fig:figure2}(a). However, these direct temporal modeling methods overlook the scene motion context, fail to achieve temporal consistency, and inherently limit the increase of effective contextual cues. Based on these limitations, we asked: \textbf{\textit{How can we accurately identify the correlation between historical frames and the current frame to guide temporal SSC modeling?}}

In this paper, we propose a novel temporal SSC method: FlowScene, Learning Temporal 3D Semantic Scene Completion via Optical Flow Guidance. As shown in Figure~\ref{fig:figure2}(b), FlowScene uses optical flow to guide temporal modeling, injecting various types of information into the SSC model, such as motion, different viewpoints, deformation, texture, geometric structure, lighting, and occlusion. As shown in Figure~\ref{fig:figure1}, the corresponding optical flow and occlusion masks are generated from the historical and current frame images, allowing for the further derivation of scene geometry and semantic structure. The positions and semantics of the car, tree trunk, vegetation, and pole within the red box in Figure~\ref{fig:figure1} are more accurate, even when they are mutually occluded.
Specifically, we introduce the \textit{Flow-Guided Temporal Aggregation} module to effectively enhance temporal and motion cues by incorporating motion and contextual information from previous frames. Furthermore, we design the \textit{Occlusion-Guided Voxel Refinement} module, which leverages aggregated features and occlusion masks to refine 3D voxel predictions for explicit geometric modeling.
To evaluate the performance of FlowScene, we conduct thorough experiments on SemanticKITTI~\cite{behley2019semantickitti} and SSCBench-KITTI360~\cite{Liao2022kitti360,li2023sscbench}. Our method achieves state-of-the-art performance. The main contributions of our work are summarized as follows:
\begin{itemize}
    \item We introduce FlowScene, a novel approach to 3D SSC that incorporates optical flow guidance to capture and model temporal and spatial dependencies across frames. 
    \item We propose the flow-guided temporal aggregation module, which effectively enhances temporal and motion cues by incorporating motion and contextual information from previous frames.
    \item We design the occlusion-guided voxel refinement module, which leverages aggregated features and occlusion masks to refine 3D voxel predictions, enabling explicit geometric modeling and improving the accuracy of scene reconstruction in occluded regions.
    \item We evaluate FlowScene on the SemanticKITTI and SSCBench-KITTI-360 benchmarks, achieving state-of-the-art performance. Our method surpasses the latest methods in both semantic and geometric analysis, demonstrating the effectiveness of optical flow-guided temporal modeling in SSC tasks.
\end{itemize}

\section{Related Work}
\paragraph{3D Semantic Scene Completion.}

The vision-based 3D Semantic Scene Completion (SSC) solution has received widespread attention in the field of autonomous driving perception. MonoScene~\cite{cao2022monoscene} was the first to infer dense 3D semantics from a single RGB image. TPVFormer~\cite{huang2023tri} introduced a tri-perspective view (TPV) representation, extending BEV with two vertical planes. OccFormer~\cite{zhang2023occformer} proposed a dual-path transformer to encode voxel features, while VoxFormer~\cite{li2023voxformer} introduced a two-stage pipeline for voxelized semantic scene understanding. SurroundOcc~\cite{wei2023surroundocc} employed 3D convolutions for progressive voxel upsampling and dense SSC ground truth generation.
OctOcc~\cite{ouyang2024octocc} utilized an octree-based representation for semantic occupancy prediction, while NDCScene~\cite{yao2023ndc} redefined spatial encoding by mapping 2D feature maps to normalized device coordinates (NDC) rather than world space.
MonoOcc~\cite{zheng2024monoocc} enhanced 3D volumetric representations using an image-conditioned cross-attention mechanism. H2GFormer~\cite{wang2024h2gformer} introduced a progressive feature reconstruction strategy to propagate 2D information across multiple viewpoints. Symphonize~\cite{jiang2024symphonize} extracted high-level instance features to serve as key-value pairs for cross-attention. HASSC~\cite{wang2024HASSC} proposed a self-distillation framework to improve the performance of VoxFormer. Stereo-based methods, such as BRGScene~\cite{li2023stereoscene}, leveraged stereo depth estimation to resolve geometric ambiguities. MixSSC~\cite{wang2025mixssc} fused forward projection sparsity with the denseness of depth-prior backward projection.
CGFormer~\cite{CGFormer} utilized a context-aware query generator to initialize context-dependent queries tailored to individual input images, effectively capturing their unique characteristics and aggregating information within the region of interest. HTCL~\cite{li2024htcl} decomposed temporal context learning into two hierarchical steps: cross-frame affinity measurement and affinity-based dynamic refinement. VLScene~\cite{wang2025vlscene} leveraged vision-language models to extract high-level semantic priors for SSC.

\paragraph{Optical Flow for Visual Perception.}
Optical flow estimation, a fundamental task in computer vision, aims to establish dense pixel-wise correspondences between consecutive frames. FlowNet~\cite{dosovitskiy2015flownet, ilg2017flownet2} introduced the first CNN-based end-to-end flow estimation pipeline, leveraging a hierarchical pyramid structure. PWC-Net~\cite{sun2018pwc} further refined this approach by incorporating multi-stage warping to handle large-displacement motion. RAFT~\cite{teed2020raft} introduced an iterative, recurrent architecture that refines residual flow predictions in a fully convolutional manner. GMFlow~\cite{xu2022gmflow} reframed optical flow as a global matching problem, directly computing feature similarities to establish correspondences.
Beyond motion estimation, optical flow has been leveraged to enhance various vision tasks. FlowTrack~\cite{zhu2018flowtrack} used optical flow to enrich feature representations and improve tracking accuracy. FGFA~\cite{zhu2017fgfa} employed flow-guided feature aggregation for end-to-end video object detection. LoSh~\cite{Yuan_2024_LoSh} utilized flow-based warping to propagate annotations across temporal neighbors, thereby boosting referring video object segmentation. DATMO~\cite{sormoli2024optical} introduced a moving object detection and tracking framework tailored for autonomous vehicles. DeVOS~\cite{Fedynyak_2024_DeVos} incorporated optical flow into scene motion modeling, using it as a prior for learnable offsets in video segmentation.

\section{Methodology}
\subsection{Preliminary}
\paragraph{Problem Setting.}
Given a set of RGB images \( I = \{I_{t-i}\}_{i=0}^{n} \), where \( n \) is the number of historical temporal images, the objective is to jointly infer the geometry and semantics of a 3D scene. This scene is represented as a voxel grid \( \mathbb{Y} \in \mathbb{R}^{X \times Y \times Z \times (M+1)} \), where \( X, Y, Z \) represent the height, width, and depth in 3D space, respectively. Each voxel in the grid is assigned a unique semantic label from the set \( C \in \{C_0, C_1, ..., C_M\} \), where \( C_0 \) represents empty space and the remaining classes \( \{C_1, ..., C_M\} \) correspond to specific semantic categories. Here, \( M \) denotes the total number of semantic classes. The goal is to learn a transformation \( \mathbb{Y} = \theta(I_{s}) \) that closely approximates the ground truth \( \hat{\mathbb{Y}} \).
\begin{figure*}[t]
\centering
  \includegraphics[width=\textwidth]{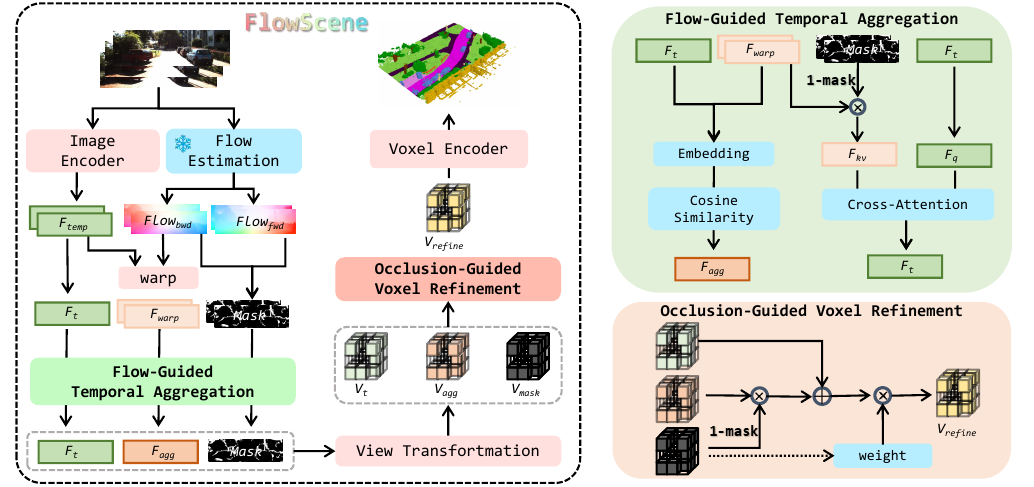}
  \caption{The FlowScene framework is proposed for temproal 3D semantic scene completion.}
  \label{fig:figure3}
\vspace{-5mm}
\end{figure*}
\subsection{Overview} 
We illustrate our method in Figure~\ref{fig:figure3}. First, we use the lightweight image encoder RepViT~\cite{wang2023repvit} and FPN~\cite{lin2017feature} to extract the current image features, $F_t$, and the historical temporal features, $F_{temp}=\{F_{t-i}\}_{i=1}^n$. We then apply the pre-trained optical flow estimation model~\cite{xu2022gmflow} (Section~\ref{sec:Flow}) to generate bidirectional optical flow, $Flow = \{Flow_{fwd}^{t-i\rightarrow t}, Flow_{bwd}^{t-i\rightarrow t}\}_{i=1}^n$. The historical temporal features, $F_{temp}$, are warped using $Flow_{bwd}$ to obtain $F_{warp} = \{F_{warp}^{t-i \rightarrow t}\}_{i=1}^n$. The bidirectional optical flow is then used for occlusion detection through a forward and backward consistency check to obtain the cumulative mask, $M \in {0,1}^{1 \times h \times w}$. Subsequently, $F_{warp}$, $F_t$, and $M$ are passed into the FGTA module (Section~\ref{sec:FGTA}) to perform optical flow-guided temporal feature aggregation in the 2D image feature space, resulting in the aggregated feature $F_{agg}$.
Next, we apply the LSS view transformation~\cite{philion2020lift} to project $F_t$, $F_{agg}$, and $M$ into the 3D voxel space, obtaining $V_t$, $V_{agg}$, and $V_{mask}$, respectively. In the subsequent OGVR module (Section~\ref{sec:OGVR}), the two voxel features are fused based on the occlusion information, yielding the refined voxel features, $V_{fine}$. Finally, $V_{fine}$ passes through the voxel encoder, then undergoes upsampling and linear projection to output the dense semantic voxels, $\mathbb{Y}$.

\subsection{Optical Flow Estimation}
\label{sec:Flow}
\paragraph{Flow-Guided Warping.}
Given a reference image frame $I_t$ and historical frames $I_{t-i}$, the flow field $Flow^{t \rightarrow t-i}=\mathcal{F}(I_t, I_{t-i})$ is estimated by a flow network $\mathcal{F}$ (e.g., GMFlow~\cite{xu2022gmflow}). The feature map from the historical frame is warped to the reference frame according to the flow. The warping function is defined as
\begin{equation}
    F_{warp}^{t-i\rightarrow t} = \mathcal{W}arp( F_{t-i},Flow^{t \rightarrow t-i})
\end{equation}
where $\mathcal{W}arp(\cdot)$ is a bilinear warping function applied to all locations of each channel in the feature map, and $F_{warp}^{t-i\rightarrow t}$ represents the feature map warped from frame $t-i$ to $t$.

\begin{figure}[t]
	\centering
	\begin{minipage}[b]{0.48\textwidth}
		\centering
		\includegraphics[width=\linewidth]{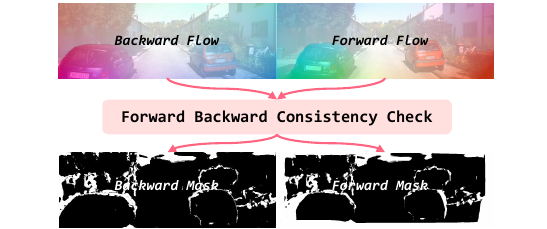}
		\caption{Occlusion detection is performed using forward-backward consistency detection.}
		\label{fig:figure4}
	\end{minipage}
	\hfill
	\begin{minipage}[b]{0.48\textwidth}
		\centering
		\includegraphics[width=\linewidth]{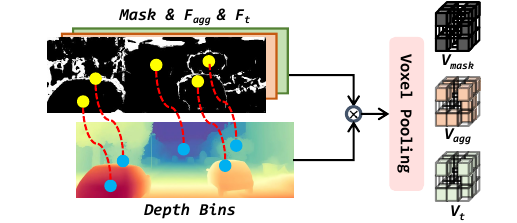}
		\caption{Projecting the occlusion mask into 3D voxel space with depth bins.}
		\label{fig:figure5}
	\end{minipage}
    \vspace{-5mm}
\end{figure}

\paragraph{Occlusion Detection.}
First, we note that there is relative motion between almost all frames in an autonomous driving scenario, which results in pixels in the current image that do not have corresponding matching pixels in the historical frames; these are referred to as occluded areas. To detect occlusion, as shown in Figure~\ref{fig:figure4}, we use the commonly employed forward and backward consistency check technique~\cite{sundaram2010dense,meister2018unflow}, which is implemented as:
\begin{equation}
    M = \mathcal{CC}(Flow^{t \rightarrow t-i},Flow^{t-i \rightarrow t}),
\end{equation}
where $\mathcal{CC}(\cdot)$ denotes the forward and backward consistency check function, and we have included a detailed explanation in the Technical Appendix. For non-occluded pixels, the forward optical flow should be the inverse of the backward optical flow of the corresponding pixel in the second frame. A pixel is marked as occluded if the mismatch between the two flows exceeds a predefined threshold. Thus, we define the occlusion flag as 1 whenever the constraint is violated and 0 otherwise.
\subsection{Flow-Guided Temporal Aggregation}
\label{sec:FGTA}
Previous SSC methods either stacked historical frame features or estimated camera poses to align features, aiming to complement the current frame. However, this direct temporal modeling approach overlooks the scene motion context, fails to achieve temporal consistency, and inherently limits the ability to leverage additional effective cues. To better incorporate time- and motion-related cues, we propose a flow-guided temporal aggregation module in 2D space. This module leverages optical flow information to align and aggregate temporal features along the motion path. As illustrated on the right side of Figure~\ref{fig:figure3}.

Specifically, guided by optical flow, the historical frame features are warped to the reference frame.
Features from different frames provide multiple information for the same object instance, such as motion, different viewpoints, deformations, textures, geometric structures, various lighting and occlusions.
First, we assign different weights to different spatial locations, while ensuring that the spatial weights remain the same across all feature channels.
At position $\textbf{P}$, if the warped feature $F_{warp}^{t-i\rightarrow t}(\text{P})$ is close to the feature $F_t(\text{P})$, it is assigned a larger weight. Otherwise, a smaller weight is assigned. Inspired by FGFA~\cite{zhu2017fgfa}, we use the cosine similarity~\cite{luo2018cosine} to measure the similarity between the warped features and the reference frame features: 
\begin{equation}
    w_{t-i\rightarrow t} (\textbf{P})= \operatorname{similarity}(F_{warp}^{t-i\rightarrow t}(\text{P}),F_t(\text{P})).
\end{equation}
Then, we use the similarity weights to aggregate these feature maps to enhance the scene motion context features. The aggregation feature ${F}_{agg}$ is obtained as:
\begin{equation}
    {F}_{agg}=\sum_{i=0}^{t}w_{t-i\rightarrow t}\cdot F^{t-i\rightarrow t}_{warp}.
\end{equation}

The non-occluded regions in the historical frames usually have richer texture and feature information, which may be missing in the current frame due to visual occlusion. To address this, we enhance the current frame features, we effectively fuse spatiotemporal information through the neighborhood cross-attention mechanism ~\cite{hassani_2023_neighborhood}.
We selectively use non-occluded features from historical frames to prevent injecting unreliable or distorted information caused by occlusions or inaccurate flow. Including occluded regions can introduce noise and harm completion quality. First, we select reliable non-occluded region features in the historical frames based on the occlusion mask. The reference features $F_t$ are used as query, and the warp features $F_{warp}$ of the non-occluded regions serve as key and value. The specific operations are as follows:
\begin{equation}
    F_t = \mathcal{NCA}(F_t,(1-M)\cdot F_{warp}),
\end{equation}
where $\mathcal{NCA}(\cdot)$ is the neighborhood cross attention mechanism. After these operations, $F_t$ fuses the non-occluded region information from both the current and historical frames, providing more stable and accurate features that enhance the perception of dynamic scenes and occluded regions.

\subsection{Occlusion-Guided Voxel Refinement} 
\label{sec:OGVR}
After passing through the FGTA module, time- and motion-related cues are injected into the image features $F_t$ and the aggregate features $F_{agg}$. However, for the 3D voxel space, there is a lack of explicit geometric modeling. To incorporate occlusion and optical flow information into the 3D space, we introduce the occlusion-guided voxel refinement module.
This module enhances the semantic completion ability of the occluded region by employing a weighted strategy of the occlusion mask. As shown in the right side of Figure~\ref{fig:figure3}.

Specifically, as shown in Figure~\ref{fig:figure5}, we follow the LSS view transformation paradigm and use depth bin $D$ assignment to project $F_t$, $F_{agg}$, and $M$ into the 3D voxel space to obtain $V_t$, $V_{agg}$, and $V_{mask}$, respectively, 
\begin{align}
   \nonumber V_t &= \operatorname{VoxelPooling}(F_t\otimes D),\\
   V_{agg} &= \operatorname{VoxelPooling}(F_{agg}\otimes D),\\
   \nonumber V_{mask} &= \operatorname{VoxelPooling}(M\otimes D),
\end{align}
where $\otimes$ is the dot product operation, VoxelPooling is the voxel pooling operation.
First, we use $V_{mask}$ to distinguish occluded and non-occluded regions in the 3D voxel space. For the non-occluded region, we use the aggregate features $V_{agg}$ that fuse multiple cues.
Subsequently, since the information from the corresponding position in the historical frame may be inaccurate due to occlusion, we use the voxel features from the current frame to update the occluded area to supplement the latest environmental information.
Finally, by constructing a weighted matrix, we normalize the fused voxel features to ensure that there is no mutation at the boundary between the occluded and non-occluded areas, thereby improving the smoothness of the features. The specific operation is as follows:
\begin{equation}
    V_{fine}=\frac{(1-V_{mask}) \cdot V_{agg} + V_t}{(1-V_{mask})+1}.
\end{equation}

Finally, $V_{fine}$ enters the sparse voxel encoder for feature extraction, and then performs linear prediction to output dense semantic voxels $\mathbb{Y}$.

\subsection{Training Loss}
In the FlowScene framework, we adopt the scene-class affinity loss $\mathcal{L}_{scal}$ from MonoScene~\cite{cao2022monoscene} to optimize precision, recall, and specificity concurrently. The scene-class affinity loss is applied to semantic and geometric predictions, in conjunction with the cross-entropy loss weighted by class frequencies. Besides, the intermediate depth distribution for view transformation is supervised by the projections of LiDAR points, with the binary cross-entropy loss $\mathcal{L}_{d}$ following BEVDepth~\cite{li2023bevdepth}. The overall loss function is formulated as follows: 
\begin{equation}
    \mathcal{L} = \lambda_{sem}\mathcal{L}^{sem}_{scal} + \lambda_{geo}\mathcal{L}^{geo}_{scal} + \lambda_{ce}\mathcal{L}_{ce} +  \lambda_{d}\mathcal{L}_{d}, 
\end{equation}
where several $\lambda$ are balancing coefficients.

\section{Experiments}
To assess the effectiveness of our FlowScene, we conducted thorough experiments using the large outdoor datasets SemanticKITTI~\cite{behley2019semantickitti,Geiger2012kitti}, SSCBench-KITTI-360~\cite{li2023sscbench,Liao2022kitti360}.
Information about datasets, metrics, and detailed implementation details is provided in the Technical Appendix, where additional experiments and analysis are also provided.

\begin{table*}[h]
  \centering

  \caption{Quantitative results on the SemanticKITTI hidden test set. The best and the second best results are in \textbf{bold} and \underline{underlined}, respectively. The “S” and “T” denote single-frame images, and temporal images, respectively.}
    \label{tab:Test Quantitative Comparison}
  \setlength{\tabcolsep}{2pt}
  \resizebox{\textwidth}{!}{
  \begin{tabular}{l|l|c|c|rrrrrrrrrrrrrrrrrrr|r}

    \toprule
    \textbf{Methods} &\textbf{Venues}&\textbf{Input}&\textbf{IoU}   
    & \rotatebox{90}{\vcenteredbox{\colorbox[RGB]{255,0,255}{\textcolor[RGB]{255,0,255}{\rule{1px}{1px}}}} \textbf{road} } 
    & \rotatebox{90}{\vcenteredbox{\colorbox[RGB]{75,0,75}{\textcolor[RGB]{75,0,75}{\rule{1px}{1px}}}} \textbf{sidewalk} } 
    & \rotatebox{90}{\vcenteredbox{\colorbox[RGB]{255,150,255}{\textcolor[RGB]{255,150,255}{\rule{1px}{1px}}}} \textbf{parking} } 
    & \rotatebox{90}{\vcenteredbox{\colorbox[RGB]{175,0,75}{\textcolor[RGB]{175,0,75}{\rule{1px}{1px}}}} \textbf{other-grnd} } 
    &  \rotatebox{90}{\vcenteredbox{\colorbox[RGB]{255,200,0}{\textcolor[RGB]{255,200,0}{\rule{1px}{1px}}}} \textbf{building} } 
    &  \rotatebox{90}{\vcenteredbox{\colorbox[RGB]{100,150,245}{\textcolor[RGB]{100,150,245}{\rule{1px}{1px}}}} \textbf{car} } 
    & \rotatebox{90}{\vcenteredbox{\colorbox[RGB]{80,30,180}{\textcolor[RGB]{80,30,180}{\rule{1px}{1px}}}} \textbf{truck} } 
    & \rotatebox{90}{\vcenteredbox{\colorbox[RGB]{100,230,245}{\textcolor[RGB]{100,230,245}{\rule{1px}{1px}}}} \textbf{bicycle} }  
    & \rotatebox{90}{\vcenteredbox{\colorbox[RGB]{30,60,150}{\textcolor[RGB]{30,60,150}{\rule{1px}{1px}}}} \textbf{motocycle} } 
    & \rotatebox{90}{\vcenteredbox{\colorbox[RGB]{0,0,255}{\textcolor[RGB]{0,0,255}{\rule{1px}{1px}}}} \textbf{other-vehicle} } 
    & \rotatebox{90}{\vcenteredbox{\colorbox[RGB]{0,175,0}{\textcolor[RGB]{0,175,0}{\rule{1px}{1px}}}} \textbf{vegetation} } 
    & \rotatebox{90}{\vcenteredbox{\colorbox[RGB]{135,60,0}{\textcolor[RGB]{135,60,0}{\rule{1px}{1px}}}}  \textbf{trunk} } 
    & \rotatebox{90}{\vcenteredbox{\colorbox[RGB]{150,240,80}{\textcolor[RGB]{150,240,80}{\rule{1px}{1px}}}} \textbf{terrain} } 
    & \rotatebox{90}{\vcenteredbox{\colorbox[RGB]{255,30,30}{\textcolor[RGB]{255,30,30}{\rule{1px}{1px}}}} \textbf{person} } 
    & \rotatebox{90}{\vcenteredbox{\colorbox[RGB]{255,40,200}{\textcolor[RGB]{255,40,200}{\rule{1px}{1px}}}} \textbf{bicylist} } 
    & \rotatebox{90}{\vcenteredbox{\colorbox[RGB]{150,30,90}{\textcolor[RGB]{150,30,90}{\rule{1px}{1px}}}}  \textbf{motorcyclist} } 
    &  \rotatebox{90}{\vcenteredbox{\colorbox[RGB]{255,120,50}{\textcolor[RGB]{255,120,50}{\rule{1px}{1px}}}} \textbf{fence} } 
    & \rotatebox{90}{\vcenteredbox{\colorbox[RGB]{255,240,150}{\textcolor[RGB]{255,240,150}{\rule{1px}{1px}}}} \textbf{pole} } 
    & \rotatebox{90}{\vcenteredbox{\colorbox[RGB]{255,0,0}{\textcolor[RGB]{255,0,0}{\rule{1px}{1px}}}} \textbf{traf.-sign} }& \textbf{mIoU}  \\
    
    \midrule
    MonoScene~\cite{cao2022monoscene}&CVPR'2022 &S& 34.16  & 54.70&27.10&24.80&5.70&14.40&18.80&3.30&0.50&0.70&4.40&14.90&2.40&19.50&1.00&1.40&0.40&11.10&3.30&2.10 & 11.08 \\
    
    TPVFormer~\cite{huang2023tri}& CVPR'2023&S& 34.25  &55.10&27.20&27.40&6.50&14.80&19.20&3.70&1.00&0.50&2.30&13.90&2.60&20.40&1.10&2.40&0.30&11.00&2.90&1.50& 11.26\\
    

    OccFormer~\cite{zhang2023occformer}& ICCV'2023&S & 34.53  & 55.90&30.30&31.50&6.50&15.70&21.60&1.20&1.50&1.70&3.20&16.80&3.90&21.30&2.20&1.10&0.20&11.90&3.80&3.70& 12.32\\


    Symphonize~\cite{jiang2024symphonize}&CVPR'2024&S&42.19&58.40&29.30&26.90&{11.70}&24.70&23.60&3.20&{3.60}&\underline{2.60}&5.60&24.20&10.00&23.10&\textbf{3.20}&1.90&\textbf{2.00}&16.10&7.70&8.00&15.04\\
    BRGScene~\cite{li2023stereoscene}&IJCAI'2024&S & 43.34 & 61.90&31.20  &30.70 & 10.70 & 24.20 & 22.80 & 2.80 & 3.40 & 2.40 & \underline{6.10} & 23.80 & 8.40 & 27.00 & 2.90 & 2.20 & 0.50 & 16.50 & 7.00 & 7.20 & 15.36\\
    CGFormer~\cite{CGFormer}&NIPS'2024&S&\underline{44.41}&\underline{64.30}&34.20&\textbf{34.10}&12.10&25.80&26.10&4.30&\underline{3.70}&1.30&2.70&24.50&\underline{11.20}&29.30&1.70&\underline{3.60}&0.40&18.70&8.70&\textbf{9.30}&16.63\\
    VoxFormer-T~\cite{li2023voxformer}& CVPR'2023&T& 43.21& 54.10& 26.90& 25.10& 7.30& 23.50& 21.70& 3.60& 1.90& 1.60& 4.10& 24.40& 8.10& 24.20& 1.60& 1.10& 0.00& 13.10& 6.60& 5.70& 13.41\\
    H2GFormer-T~\cite{wang2024h2gformer}&AAAI'2024&T&43.52&57.90&30.40&30.00&6.90&24.00&23.70&5.20&0.60&1.20&5.00&25.20&10.70&25.80&1.10&0.10&0.00&14.60&7.50&\textbf{9.30}&14.60 \\
    HASSC-T~\cite{wang2024HASSC}&CVPR'2024&T&42.87&55.30&29.60&25.90&11.30&23.10&23.00&2.90&1.90&1.50&4.90&24.80&9.80&26.50&1.40&3.00&0.00&14.30&7.00&7.10&14.38 \\
    SGN~\cite{mei2024sgn}&TIP'2024&T&43.71&57.90&29.70&25.60&5.50&\underline{27.00}&25.00&1.50&0.90&0.70&3.60&\textbf{26.90}&\textbf{12.00}&26.40&0.60&0.30&0.00&14.70 &9.00& 6.40& 14.39\\
    HTCL~\cite{li2024htcl}&ECCV'2024&T&44.23&\textbf{64.40}&\underline{34.80}&\textbf{33.80}&\underline{12.40}&25.90&\textbf{27.30}&\underline{5.70}&1.80&2.20&5.40&25.30&10.80&\textbf{31.20}&1.10&3.10&0.90&\textbf{21.10}&\textbf{9.00}&8.30&\underline{17.09} \\
    \hline

    \rowcolor{gray!20}\textbf{Ours}&  &T& \textbf{45.20} &64.10&\textbf{35.00}&{33.70}&\textbf{13.00}&\textbf{27.70}&\underline{26.40}&\textbf{10.00}&\textbf{4.20}&\textbf{3.10}&\textbf{7.00}&\underline{26.30}&10.00&\underline{30.20}&\underline{3.10}&\textbf{5.10}&\underline{1.10}&\underline{20.20}&\underline{8.90}&\underline{9.10}& \textbf{17.70}\\
    \bottomrule
  \end{tabular}}
  \vspace{-5mm}
\end{table*}
\subsection{Main Results}
\paragraph{Quantitative Results.}
As shown in Table~\ref{tab:Test Quantitative Comparison}, we compare FlowScene with the latest public methods on the SemanticKITTI dataset, including approaches that use single-image input (S) and temporal image input (T). Temporal methods, such as VoxFormer~\cite{li2023voxformer}, H2GFormer~\cite{wang2024h2gformer}, HASSC~\cite{wang2024HASSC}, and SGN~\cite{mei2024sgn}, utilize additional historical 5-frame input, while HTCL~\cite{li2024htcl} uses a 3-frame historical input. In contrast, FlowScene uses only 2 historical frames as input, achieving the highest mIoU for the overall semantic metric and the highest IoU for the completion metric.
Compared to the best-performing HTCL with temporal input, FlowScene improves the mIoU and IoU by 0.61\% and 0.97\%, respectively. When compared to the best CGFormer, which uses single-frame input, FlowScene achieves improvements of 1.07\% in mIoU and 0.79\% in IoU. 
Additionally, our method achieves the best or second-best results in most categories, outperforming or closely matching other methods. These results demonstrate the superiority of FlowScene in both geometry and semantics, effectively utilizing optical flow motion information and achieving temporal consistency.
\begin{table*}[t]
  \centering
      \caption{Quantitative results on the SSCBench-KITTI360 test set. The best and the second best results are in \textbf{bold} and \underline{underlined}, respectively.}
  \setlength{\tabcolsep}{2pt}
  \resizebox{\textwidth}{!}{
  \begin{tabular}{l|ccc|rrrrrrrrrrrrrrrrrr|r}
    \toprule
    \textbf{Methods}&\textbf{Prec.}&\textbf{Rec.} &\textbf{IoU}   
    &  \rotatebox{90}{\vcenteredbox{\colorbox[RGB]{100,150,245}{\textcolor[RGB]{100,150,245}{\rule{1px}{1px}}}} \textbf{car} } 
    & \rotatebox{90}{\vcenteredbox{\colorbox[RGB]{100,230,245}{\textcolor[RGB]{100,230,245}{\rule{1px}{1px}}}} \textbf{bicycle} }  
    & \rotatebox{90}{\vcenteredbox{\colorbox[RGB]{30,60,150}{\textcolor[RGB]{30,60,150}{\rule{1px}{1px}}}} \textbf{motocycle} } 
    & \rotatebox{90}{\vcenteredbox{\colorbox[RGB]{80,30,180}{\textcolor[RGB]{80,30,180}{\rule{1px}{1px}}}} \textbf{truck} } 
    & \rotatebox{90}{\vcenteredbox{\colorbox[RGB]{0,0,255}{\textcolor[RGB]{0,0,255}{\rule{1px}{1px}}}} \textbf{other-vehicle} } 
    & \rotatebox{90}{\vcenteredbox{\colorbox[RGB]{255,30,30}{\textcolor[RGB]{255,30,30}{\rule{1px}{1px}}}} \textbf{person} } 
    &  \rotatebox{90}{\vcenteredbox{\colorbox[RGB]{255,0,255}{\textcolor[RGB]{255,0,255}{\rule{1px}{1px}}}} \textbf{road} } 
    & \rotatebox{90}{\vcenteredbox{\colorbox[RGB]{255,150,255}{\textcolor[RGB]{255,150,255}{\rule{1px}{1px}}}} \textbf{parking} } 
    & \rotatebox{90}{\vcenteredbox{\colorbox[RGB]{75,0,75}{\textcolor[RGB]{75,0,75}{\rule{1px}{1px}}}} \textbf{sidewalk} } 
    & \rotatebox{90}{\vcenteredbox{\colorbox[RGB]{175,0,75}{\textcolor[RGB]{175,0,75}{\rule{1px}{1px}}}} \textbf{other-grnd} } 
    &  \rotatebox{90}{\vcenteredbox{\colorbox[RGB]{255,200,0}{\textcolor[RGB]{255,200,0}{\rule{1px}{1px}}}} \textbf{building} } 
    &  \rotatebox{90}{\vcenteredbox{\colorbox[RGB]{255,120,50}{\textcolor[RGB]{255,120,50}{\rule{1px}{1px}}}} \textbf{fence} } 
    & \rotatebox{90}{\vcenteredbox{\colorbox[RGB]{0,175,0}{\textcolor[RGB]{0,175,0}{\rule{1px}{1px}}}} \textbf{vegetation} } 
    & \rotatebox{90}{\vcenteredbox{\colorbox[RGB]{150,240,80}{\textcolor[RGB]{150,240,80}{\rule{1px}{1px}}}} \textbf{terrain} } 
    & \rotatebox{90}{\vcenteredbox{\colorbox[RGB]{255,240,150}{\textcolor[RGB]{255,240,150}{\rule{1px}{1px}}}} \textbf{pole} } 
    & \rotatebox{90}{\vcenteredbox{\colorbox[RGB]{255,0,0}{\textcolor[RGB]{255,0,0}{\rule{1px}{1px}}}} \textbf{traf.-sign} } 
    & \rotatebox{90}{\vcenteredbox{\colorbox[RGB]{0, 150, 255}{\textcolor[RGB]{0, 150, 255}{\rule{1px}{1px}}}}  \textbf{other-struct.} } 
    & \rotatebox{90}{\vcenteredbox{\colorbox[RGB]{255, 255, 50}{\textcolor[RGB]{255, 255, 50}{\rule{1px}{1px}}}} \textbf{other-obj.} } &\textbf{mIoU}   \\
    
    \midrule
    MonoScene&56.73 &53.26&37.87&19.34&0.43&0.58&8.02&2.03&0.86&48.35&11.38&28.13&3.32&32.89&3.53&26.15&16.75&6.92&5.67&4.20&3.09&12.31\\
    VoxFormer&58.52 &53.44&38.76&17.84&1.16&0.89&4.56&2.06&1.63&47.01&9.67&27.21&2.89&31.18&4.97&28.99&14.69&6.51&6.92&3.79&2.43&11.91\\
    TPVFormer&59.32 &\underline{55.54}&40.22&21.56&1.09&1.37&8.06&2.57&2.38&52.99&11.99&31.07& 3.78&34.83&4.80&30.08&17.52&7.46&5.86&5.48&2.70&13.64\\
    OccFormer&59.70 &55.31&40.27&22.58&0.66&0.26&9.89&3.82&2.77&54.30&13.44&31.53&3.55&36.42&4.80&31.00&\underline{19.51}&7.77&8.51& 6.95&4.60&13.81\\
    IAMSSC&-&-&41.80&18.53&\underline{2.45}&1.76&5.12&3.92&3.09&47.55&10.56&28.35&4.12&31.53&6.28&29.17&15.24&8.29&7.01&6.35&4.10&12.97\\
    Symphonies&\underline{69.24}&54.88&44.12&\underline{30.02}&1.85&\textbf{5.90}&\underline{25.07}&\underline{12.06}&\underline{8.20}&54.94&13.83&32.76&\underline{6.93}&35.11&\underline{8.58}&{38.33}&11.52&14.01&9.57&\underline{14.44}&\underline{11.28}&18.58\\
    CGFormer&-&-&\textbf{48.07}&29.85&\textbf{3.42}&3.96&17.59&6.79&6.63&\textbf{63.85}&\textbf{17.15}&\textbf{40.72}&5.53&\textbf{42.73}&8.22&\underline{38.80}&\textbf{24.94}&\underline{16.24}&\textbf{17.45}&10.18&6.77&\underline{20.05}\\
    
    \hline
    \rowcolor{gray!20}\textbf{Ours}& \textbf{69.70} &\textbf{58.99} &\underline{46.95} &\textbf{32.48}&{1.87}&\underline{4.93}&\textbf{25.47}&\textbf{14.86}&\textbf{9.62}&\underline{60.53}&\underline{16.49}&\underline{36.13}&\textbf{8.58}&\underline{39.66}&\textbf{9.62}&\textbf{39.82}&{13.32}&\textbf{17.52}&\underline{14.35}&\textbf{16.25}&\textbf{13.08}& \textbf{20.81}\\
    \bottomrule
  \end{tabular}}
\vspace{-5mm}

  \label{tab:KITTI360Test Quantitative Comparison}
\end{table*}

To demonstrate the diversity of our model, we conducted experiments on the SSCBench-KITTI-360 dataset, as shown in Table~\ref{tab:KITTI360Test Quantitative Comparison}. It is worth noting that our method has a huge advantage in the performance of potential moving objects (such as car, truck, other-vehicle, person, etc.)

Moreover, Table~\ref{tab:diff range} illustrates the performance of FlowScene across three distance ranges (12.5m, 25.6m, 51.2m) on the SemanticKITTI validation set. It is evident that our approach significantly outperforms state-of-the-art methods at every tested distance.
Furthermore, as shown in Table~\ref{tab:cost}, we compare the inference time and number of parameters of our method with other state-of-the-art methods on the SemanticKITTI validation set. The inference time of MonoScene is optimal because of its FLoSP feature projection method. But, FlowScene achieves state-of-the-art performance with a mIoU of 18.13\%, while utilizing only 52.4M parameters. Additionally, FlowScene processes the extra 2-frame temporal image input with lower inference time, further demonstrating its efficiency and superior mIoU performance.
\begin{center}
	\begin{minipage}[t]{0.48\textwidth} 
		\centering
		\renewcommand\arraystretch{1}
		\small
		\captionof{table}{Comparison of different ranges on SemanticKITTI validation set.} 
		\label{tab:diff range}
		\resizebox{\linewidth}{!}{
		  \begin{tabular}{l|l|ccc}
			\toprule
			\multirow{2}{*}{\textbf{Methods}}  & \multirow{2}{*}{\textbf{Venues}}& & \textbf{mIoU(\%)} &  \\
			& &12.8m &25.6m & 51.2m  \\
			\midrule
			MonoScene  & CVPR'2022&12.25& 12.22& 11.30\\
			VoxFormer-T& CVPR'2023 & 21.55& 18.42 &13.35\\
			HASSC-T& CVPR'2024& 24.10 &20.27 &14.74\\
			H2GFormer-T& AAAI'2024 & 23.43& 20.37& 14.29 \\
			BRGScene& IJCAI'2024 & 23.27& 21.15& 15.24\\
			SGN-T&TIP'2024&25.70 &22.02& 15.32\\
            VLScene&AAAI'2025&26.51& 24.37& 17.83\\
			\rowcolor{gray!20}\textbf{Ours} & &  \textbf{27.63}& \textbf{24.65}& \textbf{18.13}\\
			\bottomrule
		\end{tabular} 
			}
	\end{minipage}
	\hfill 
	\begin{minipage}[t]{0.48\textwidth}
		\centering
		\renewcommand\arraystretch{1}
		\small
		\captionof{table}{Comparison of inference time and number of parameters.}
		\label{tab:cost}
		\resizebox{\linewidth}{!}{
		  \begin{tabular}{l|c|ccc}
			\toprule
			{\textbf{Method}}&\textbf{Input}& {\textbf{mIoU}(\%)}&\textbf{Times(s)}&\textbf{Params(M)} \\
			\midrule
			MonoScene& T&12.96&\textbf{0.281}&132.4\\
			OccFormer&T& {13.58}&0.338&203.4 \\
			VoxFormer&T& {13.35}&0.307&57.9 \\
			Symphonize&S& {14.89}&0.319&59.3 \\
			BRGScene&S& {15.43}&0.285&161.4 \\
			HTCL&T& {17.13}&0.297&181.4 \\
			\rowcolor{gray!20}{\textbf{Ours}}&T&{\textbf{18.13}}&0.301&\textbf{52.4} \\
			\bottomrule
		\end{tabular}
	}
	\end{minipage}
    \vspace{-3mm}
    
\end{center}

\begin{figure*}[t]
\centering
  \includegraphics[width=0.95\textwidth]{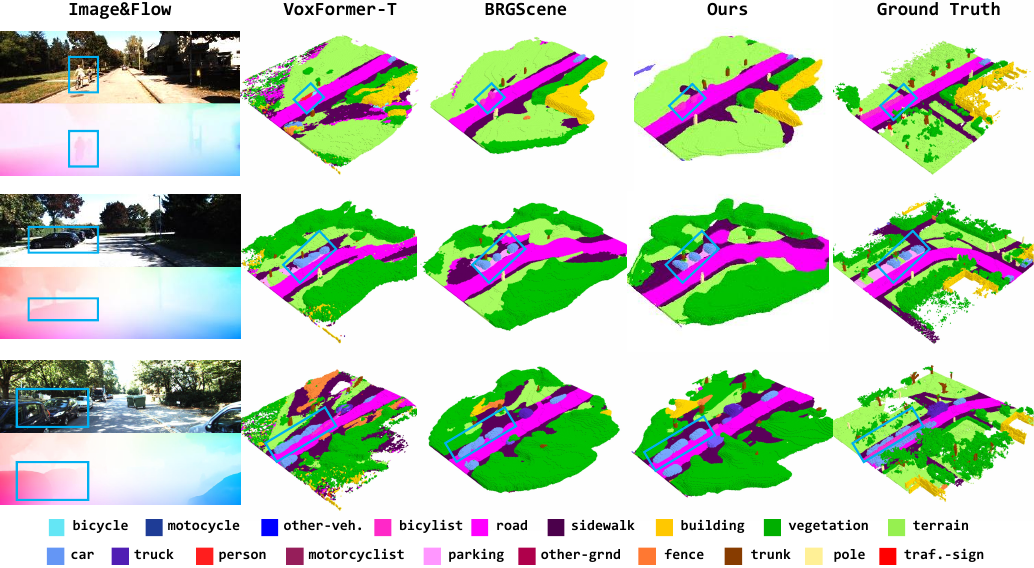}
  \caption{Qualitative results on the SemanticKITTI validation set.}
  \label{fig:figure6}
\end{figure*}

\paragraph{Qualitative Visualizations.}
To intuitively demonstrate the performance of FlowScene, Figure~\ref{fig:figure6} presents qualitative results for VoxFormer-T, BRGScene, and our method on the SemanticKITTI validation set.
The first column displays the input reference image and the corresponding optical flow. It is evident that optical flow is particularly sensitive to the perception of moving objects, such as cars and cyclists.
Compared to BRGScene, our method more effectively captures the location and details of mutually occluded objects in the scene (e.g., the arrangement of multiple cars in the second row).
In comparison to VoxFormer-T, FlowScene maintains better temporal consistency, as shown by the car parked on the roadside in the blue box in the third row.
Overall, our method demonstrates superior geometric and semantic visualization.

\begin{table*}[h]
  \centering
  \caption{Analysis of Static and Dynamic Objects}
  \label{tab:dyna}
  \resizebox{0.75\textwidth}{!}{
  \begin{tabular}{c|ccc|ccc}
    \toprule
    \multirow{2}{*}{\textbf{Method}} & \multicolumn{3}{c}{{SemanticKITTI}} & \multicolumn{3}{c}{SSCBench-KITTI-360}  \\
      & {Dynamic} &{Static}&All-mIoU&{Dynamic}&{Static}&All-mIoU\\
    \midrule
    MonoScene& 3.81&16.36&11.08&5.21&15.87&12.13\\
    VoxFormer& 4.45&19.91&13.41&4.69&12.19&11.91\\
    Symphonie& 5.71&21.83&15.04&13.85&20.94&18.58\\
    CGFormer& 5.48&24.75&16.63&11.37&\textbf{24.38}&20.05\\
    \textbf{Ours}& \textbf{7.50}&\textbf{25.29}&\textbf{17.70}&\textbf{14.87}&23.78&\textbf{20.81}\\
    \bottomrule
  \end{tabular}}
\end{table*}
\begin{figure*}[h]
\centering
  \includegraphics[width=0.75\textwidth]{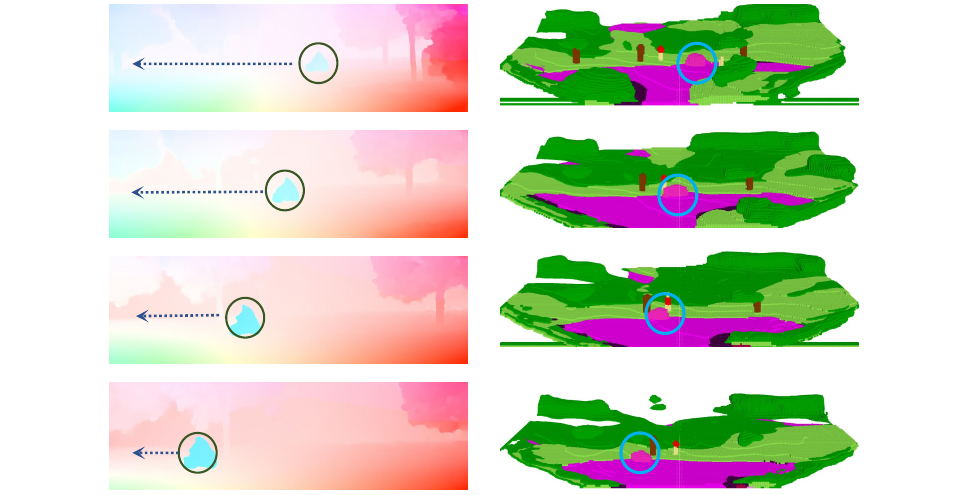}
  \caption{Dynamic object modeling case visualization results.}
  \label{fig:sup4}
\end{figure*}
\subsection{Analysis of Static and Dynamic Objects}
\label{static-dynamic}
To better understand the performance of our method across different types of semantic categories, we conduct a detailed analysis by separating static and dynamic object classes in the autonomous driving datasets.
For SemanticKITTI, we define dynamic objects as the classes that are likely to involve motion: \vcenteredbox{\colorbox[RGB]{100,150,245}{\textcolor[RGB]{100,150,245}{\rule{1px}{1px}}}} car, \vcenteredbox{\colorbox[RGB]{80,30,180}{\textcolor[RGB]{80,30,180}{\rule{1px}{1px}}}} truck, \vcenteredbox{\colorbox[RGB]{100,230,245}{\textcolor[RGB]{100,230,245}{\rule{1px}{1px}}}} bicycle, \vcenteredbox{\colorbox[RGB]{30,60,150}{\textcolor[RGB]{30,60,150}{\rule{1px}{1px}}}} motocycle, \vcenteredbox{\colorbox[RGB]{0,0,255}{\textcolor[RGB]{0,0,255}{\rule{1px}{1px}}}} other-vehicle, \vcenteredbox{\colorbox[RGB]{255,30,30}{\textcolor[RGB]{255,30,30}{\rule{1px}{1px}}}} person and \vcenteredbox{\colorbox[RGB]{255,40,200}{\textcolor[RGB]{255,40,200}{\rule{1px}{1px}}}} bicylist as dynamic objects and others as static objects. For SSCBench-KITTI-360, we classified \vcenteredbox{\colorbox[RGB]{100,150,245}{\textcolor[RGB]{100,150,245}{\rule{1px}{1px}}}} car, \vcenteredbox{\colorbox[RGB]{100,230,245}{\textcolor[RGB]{100,230,245}{\rule{1px}{1px}}}} bicycle, \vcenteredbox{\colorbox[RGB]{30,60,150}{\textcolor[RGB]{30,60,150}{\rule{1px}{1px}}}} motocycle, \vcenteredbox{\colorbox[RGB]{80,30,180}{\textcolor[RGB]{80,30,180}{\rule{1px}{1px}}}} truck, \vcenteredbox{\colorbox[RGB]{0,0,255}{\textcolor[RGB]{0,0,255}{\rule{1px}{1px}}}} other-vehicle and \vcenteredbox{\colorbox[RGB]{255,30,30}{\textcolor[RGB]{255,30,30}{\rule{1px}{1px}}}} person as dynamic objects and others as static objects. 
Table~\ref{tab:dyna} presents a comparative evaluation of our method against several prior state-of-the-art approaches, reporting mean IoU (mIoU) separately for dynamic objects, static objects, and the overall average.

Our method clearly outperforms all baselines in both datasets. Notably:
On SemanticKITTI, our model achieves 7.50\% mIoU for dynamic objects—the highest among all methods, surpassing CGFormer (5.48\%) and Symphonie (5.71\%). For static objects, we also lead with 25.29\%, again outperforming CGFormer (24.75\%).
On SSCBench-KITTI-360, our method obtains 14.87\% mIoU on dynamic objects, significantly ahead of CGFormer (11.37\%) and Symphonie (13.85\%). Although CGFormer slightly surpasses us on static objects (24.38\% vs. 23.78\%), our method achieves the highest overall score of 20.81\% mIoU.
These results demonstrate that FlowScene plays a crucial role in handling motion and temporal variation, making it especially effective in recognizing and completing dynamic objects. Unlike previous approaches that rely on frame stacking or static assumptions, our model adapts to motion patterns, improving semantic consistency across time.

Figure~\ref{fig:sup4} illustrates the qualitative results of dynamic object modeling in a challenging real-world scenario. As shown, the cyclist in motion is effectively captured by the optical flow module, which accurately estimates the movement across frames. By leveraging this motion information, our model performs kinematic compensation, aligning temporal features and preserving object structure throughout the sequence. This results in a more coherent and complete 3D semantic reconstruction of the dynamic scene.
To further support this case, we provide a supplementary video that offers an intuitive, frame-by-frame visualization of the temporal alignment and dynamic object modeling. This additional material highlights the superior capability of our method to handle complex motion compared to static-assumption baselines.

\begin{table*}[t]
\centering\small
  \caption{Ablation study for Architecture Components on SemanticKITTI validation set. OFE: Optical Flow Estimation; FGTA: Flow-Guided Temporal Aggregation; OGVR: Occlusion-Guided Voxel Refinement; FGW: Flow-Guided Warping; OD: Occlusion Detection; TA: Temporal Aggregation; OCA: Occlusion Cross-Attention; $\operatorname{V_t}$: reference voxel features; $ \operatorname{V_{agg}}$: aggregation voxel features; $ \operatorname{V_{mask}}$: voxel occlusion mask.}
\resizebox{0.75\textwidth}{!}{
  \begin{tabular}{c|cc|cc|ccc|ccc}
    \toprule
     \multirow{2}{*}{\textbf{Variants}}& \multicolumn{2}{c|}{\textbf{OFE}} &\multicolumn{2}{c|}{\textbf{FGTA}} &   \multicolumn{3}{c|}{\textbf{OGVR}} & \multirow{2}{*}{\textbf{IoU(\%)}} & \multirow{2}{*}{\textbf{mIoU(\%)}}& \multirow{2}{*}{\textbf{Params(M)}} \\
     &FGW&OD&TA&{OCA}&$\operatorname{V_t}$ &$\operatorname{V_{agg}}$ &$\operatorname{V_{mask}}$&&&\\

    \midrule
      Baseline&{}& {}&{}& {} & {} & {} &{} &  {43.98} & {15.89}&47.4\\
      
     1&{\checkmark}& {}&{}& {} & {} & {} &{} &  {44.13} & {16.21}&52.1 \\
     
     2&{\checkmark}& {\checkmark}&{}& {\checkmark} & {\checkmark} & {} &{} &  {44.38} & {16.43}&52.2 \\
     
     3&{\checkmark}& {\checkmark}&{}& {} & {\checkmark} & {\checkmark} &{\checkmark} &  {44.56} & {16.67}&52.1 \\
     
     4&{\checkmark}& {\checkmark} & {\checkmark}&{}& {\checkmark} & {\checkmark} &{\checkmark} &  {44.63} & {17.23}&52.3 \\
     
     5&{\checkmark}& {\checkmark}&{}& {\checkmark} & {\checkmark} & {\checkmark} &{\checkmark} &  {44.42} & {17.08}&52.3 \\
     
     6&{\checkmark}& {\checkmark} & {\checkmark}&{\checkmark}& {\checkmark} & {} & {} & {44.68} & {17.18}&52.4 \\
     
     7&{\checkmark}&{\checkmark }&{\checkmark}& {\checkmark} & {\checkmark} & {\checkmark} &{}& {{44.72}} &{{17.63}}&{52.4} \\
     
     \rowcolor{gray!20}8&{\checkmark}&{\checkmark }&{\checkmark}& {\checkmark} & {\checkmark} & {\checkmark} &{\checkmark}& {\textbf{45.01}} &{\textbf{18.13}}&{52.4} \\
    \bottomrule

\end{tabular}}

  \label{tab:architecture}
     \vspace{-5mm}
\end{table*}

\subsection{Ablation Studies}
We conduct extensive ablation experiments for FlowScene on the Semantickitti validation set. Specifically, we analyze the impact of different architecture component variations in Table~\ref{tab:architecture}.
\vspace{-3mm}
\paragraph{Optical Flow Estimation (OFE).}
The baseline model removes all components, using only the current image and two frames of historical images as input. After passing through the image encoder, all features are stacked together. Variant 1 in Table~\ref{tab:architecture} uses Flow-Guided Warping to align the temporal features to the reference moment, achieving a 0.32\% mIoU improvement (Variant 1 vs. Baseline).
Additionally, Variant 2 incorporates Occlusion Detection to obtain an occlusion mask, which guides the interaction of non-occluded areas in the 2D feature space, boosting the mIoU score by 0.22\% (Variant 2 vs. Variant 1).
\vspace{-3mm}

\paragraph{Flow-Guided Temporal Aggregation (FGTA).}
In Table~\ref{tab:architecture}, Variants 3, 4, and 5 represent different configurations of the FGTA module: removing the FGTA module (Variant 3), removing the Occlusion Cross-Attention (Variant 4), and removing the Temporal Aggregation component (Variant 5). Variant 4 adaptively assigns weights to aggregate historical features, resulting in a 0.56\% mIoU improvement (Variant 4 vs. Variant 3). Variant 5 uses Occlusion Cross-Attention to facilitate interaction between the current feature and the non-occluded areas in the historical frame, enhancing the texture and contextual information of the current frame's features, further boosting the mIoU by 0.41\% (Variant 5 vs. Variant 3).
\vspace{-3mm}

\paragraph{Occlusion-Guided Voxel Refinement (OGVR).}
In Table~\ref{tab:architecture}, Variant 6 represents the removal of the OGVR module, while Variant 7 uses convolution fusion to concatenate $V_t$ and $V_{agg}$. Even with this simple fusion strategy, a 0.45\% mIoU improvement is achieved (Variant 7 vs. Variant 6). Variant 8 represents our final full model. Compared to Variant 7, the mask-based refinement strategy further improves the mIoU metric. It is worth noting that the OGVR module incurs no additional parameter overhead. 
\begin{center}
	\begin{minipage}[t]{0.4\textwidth} 
		\centering
		\renewcommand\arraystretch{1}
		\small
		\captionof{table}{Ablation study for temporal alignment strategy.} 
        \label{table:strategy}
		\resizebox{\linewidth}{!}{
  \begin{tabular}{l|cc}
    \toprule
    {\textbf{Method}}& {\textbf{IoU}(\%)}&{\textbf{mIoU}(\%)} \\
    \midrule
    Stack& 43.98&15.89\\
VoxFormer-T~\cite{li2023voxformer}&44.15&13.35\\
    HTCL~\cite{li2024htcl}&\textbf{45.51}&17.13\\
    \rowcolor{gray!20}{Flow-Guided [Ours]}&{{45.01}}&\textbf{18.13} \\
    \bottomrule
  \end{tabular}
		}
	\end{minipage}
	\hfill 
	\begin{minipage}[t]{0.58\textwidth}
		\centering
		\renewcommand\arraystretch{1}
		\small
		\captionof{table}{Ablation study for different number of temporal inputs.}
        \label{table:num}
        
		\resizebox{\linewidth}{!}{
  \begin{tabular}{ccccc|ccc}
    \toprule
    \multicolumn{5}{c|}{\textbf{ Temporal Inputs}}&\multirow{2}{*}{\textbf{IoU(\%)}}  & \multirow{2}{*}{\textbf{mIoU(\%)}} &\multirow{2}{*}{\textbf{Times(s)}}\\
    t-1&t-2 &t-3 &t-4 &t-5&&  \\
    \midrule
    {\checkmark}& {} & {}&{}& {} & {44.63} & {17.74}&0.290\\
    \rowcolor{gray!20}{\checkmark}& {\checkmark} & {}&{}& {} & {45.01} & {18.13}& {0.301}\\
    {\checkmark}& {\checkmark} & {\checkmark}&{}& {} & {44.72} & {18.30}& {0.314}\\
    {\checkmark}& {\checkmark} & {\checkmark}&{\checkmark}& {} & {44.66} & {17.68}& {0.328}\\
    {\checkmark}& {\checkmark} & {\checkmark}&{\checkmark}& {\checkmark} & {44.53} & {17.55}& {0.344}\\
    \bottomrule
  \end{tabular} 
		}
	\end{minipage}
\end{center}

\begin{center}
	\begin{minipage}[t]{0.40\textwidth} 
		\centering
		\renewcommand\arraystretch{1}
		\small
		\captionof{table}{Ablation study for optical flow networks.} 
        \label{table:optical}
		\resizebox{\linewidth}{!}{
  \begin{tabular}{l|ccc}
	\toprule
	{\textbf{Method}}& {\textbf{IoU}(\%)}&{\textbf{mIoU}(\%)}&\textbf{Params(M)} \\
	\midrule
	PWC-Net+~\cite{sun2018pwc}& 43.31&17.13&8.8\\
	RAFT~\cite{teed2020raft}&44.12&17.56&5.3 \\
	FlowFormer~\cite{huang2022flowformer}&44.33&17.74&18.2 \\
	\rowcolor{gray!20}{GMFlow}~\cite{xu2022gmflow}&{\textbf{45.01}}&\textbf{18.13}&\textbf{4.7} \\
	\bottomrule
\end{tabular}
		}
	\end{minipage}
	\hfill 
	\begin{minipage}[t]{0.58\textwidth}
		\centering
		\renewcommand\arraystretch{1}
		\small
		\captionof{table}{Ablation study for image backbone networks. Acc represents the classification accuracy of each pre-trained model on ImageNet~\cite{russakovsky2015imagenet}.}
        \label{table:backbone}
		\resizebox{\linewidth}{!}{
  \begin{tabular}{l|cccc}
	\toprule
	{\textbf{Method}}& {\textbf{IoU}(\%)}&{\textbf{mIoU}(\%)}&\textbf{Params(M)} &\textbf{Acc}(\%)\\
	\midrule

	ResNet50~\cite{he2016resnet}&44.12&16.98&25.6&79.3 \\
    EfficientNetB7~\cite{tan2019efficientnet}& 44.31&17.63&63.8&\textbf{84.4}\\
	\rowcolor{gray!20}{RepVit-M2.3}~\cite{wang2023repvit}&{\textbf{45.01}}&\textbf{18.13}&\textbf{22.4}&83.3 \\
	\bottomrule
\end{tabular}
		}
	\end{minipage}
\end{center}
Overall, compared to the baseline, our method achieves significant improvements in both completion and semantic metrics (+2.24\% mIoU, +1.03\% IoU).
\vspace{-3mm}
\paragraph{Temporal Alignment Strategy and Temporal Inputs.}
Table~\ref{table:strategy} presents the results of an ablation study comparing different temporal alignment strategies used for fusing features across frames. Our Flow-Guided strategy delivers the best mIoU of 18.13\%, showing that optical flow-guided alignment is highly effective for preserving fine-grained semantic consistency, particularly for dynamic scenes.
As shown in Table~\ref{table:num}, we evaluate the performance of temporal inputs with different numbers of frames. We observe that, as the number of frames increases, the time overhead also increases. However, the mIoU metric does not grow linearly, as the quality of optical flow prediction decreases when the time interval between frames is longer. As a result, inputs with 4 or 5 frames (t-4 and t-5) lead to reduced effectiveness. Considering both the experimental metrics and the time overhead, we use 2 frames as the input for our FlowScene method.

\vspace{-3mm}
\paragraph{Optical Flow Networks.}
Table~\ref{table:optical} presents the performance of different optical flow networks. We compare several state-of-the-art methods, including PWC-Net~\cite{sun2018pwc}, RAFT~\cite{teed2020raft}, and FlowFormer~\cite{huang2022flowformer}, along with our setting, GMFlow~\cite{xu2022gmflow}, which is highlighted in the last row.
Our setting achieves the highest IoU of 45.01\% and mIoU of 18.13\%, outperforming all other methods in both metrics. These results suggest that GMFlow effectively captures motion cues and integrates them into the semantic scene completion task, providing superior performance over the other optical flow networks tested, with significantly fewer parameters.
\vspace{-3mm}
\paragraph{Image Backbone Networks.}

Table~\ref{table:backbone} examines the impact of different backbone networks on the performance of FlowScene. The study compares EfficientNetB7~\cite{tan2019efficientnet}, ResNet50~\cite{he2016resnet}, and RepVit-M2.3~\cite{wang2023repvit}(our setting). Our method, using RepVit-M2.3, achieves the highest IoU of 45.01\% and mIoU of 18.13\%, surpassing both EfficientNetB7 (44.31\% IoU, 17.63\% mIoU) and ResNet50 (44.12\% IoU, 16.98\% mIoU). RepVit-M2.3, though achieving the best performance, maintains a relatively low parameter count of 22.4M. In comparison, EfficientNetB7 has a much higher parameter count of 63.8M, while ResNet50 is more parameter-efficient at 25.6M. RepVit-M2.3 offers a good balance between performance and parameter count, making it an ideal choice for our backbone network. Different image encoders have significant improvements over the baseline, demonstrating the effectiveness of our entire model.
\vspace{-3mm}
\section{Conclusion}
In this paper, we propose a novel temporal SSC method FlowScene. Specifically, we introduce a Flow-Guided Temporal Aggregation module that aligns and aggregates temporal features using optical flow, capturing motion-aware context and deformable structures. In addition, we design an Occlusion-Guided Voxel Refinement module that injects occlusion masks and temporally aggregated features into 3D voxel space, adaptively refining voxel representations for explicit geometric modeling.
Experimental results demonstrate that FlowScene achieves SOTA performance on the SemanticKITTI and SSCBench-KITTI-360 benchmarks.

\begin{ack}
This work was supported in part by the State Key Laboratory of Advanced Design and Manufacturing Technology for Vehicle under Grant 72365001, in part by the National Natural Science Foundation of China(Grant Nos. 62225205, 62472162, 62473137), in part by the National Key Researchand Development Program of China (No.2025YFB3003601), in part by the Major Science and Technology Research Projects of Hunan Province (Grant Nos. 2024QK2010, 2024QK2009).
\end{ack}

\bibliography{main}
\bibliographystyle{plain}





\clearpage

\appendix

\section*{Technical Appendices and Supplementary Material}
This technical appendices consists of the following sections:
\begin{itemize}
\item
In Section~\ref{exp}, we provide information of datasets, metrics and detailed implementation details.
\item
In Section~\ref{results}, we provide more experiments results and analysis.
\item
In Section~\ref{sec:limit}, we analyze the limitations of our approach, directions for future work, and the broader impacts of this work.
\end{itemize} 
We also include a video in the supplementary material.
\section{Experimental Setup}
\label{exp}
\paragraph{Datasets.} 
The SemanticKITTI\cite{behley2019semantickitti, Geiger2012kitti} dataset includes dense semantic scene completion annotations and labels a voxelized scene with 20 semantic classes. It consists of 10 training sequences, 1 validation sequence, and 11 testing sequences. RGB images are resized to $1280\times384$ for input processing.
The SSCBench-KITTI-360\cite{li2023sscbench, Liao2022kitti360} dataset contains 7 training sequences, 1 validation sequence, and 1 testing sequence, covering 19 semantic classes in total. The RGB images are resized to $1408\times384$ for input processing.

\paragraph{Metrics.} We use intersection over union (IoU) to evaluate the scene completion performance. To assess the effectiveness of our 3D Semantic Scene Completion method, we focus on the mean IoU (mIoU). A higher IoU value reflects accurate geometric predictions, while a higher mIoU value indicates more precise semantic segmentation.

\paragraph{Training  Details.} 
We use RepVit~\cite{wang2023repvit} and FPN~\cite{lin2017feature} to extract features for all images. The number of historical temporal frames n is set to 2.
We use and freeze the GMFlow~\cite{xu2022gmflow} optical flow estimation model to obtain optical flow information.
We use the LSS paradigm for 2D-3D projection. The neighborhood cross-attention range is set to 7, and the number of attention heads is set to 8. Finally, the final outputs of SemantiKITTI is 20 classes, and SSCBench-KITTI-360 is 19 classes. All datasets have the scene size of $51.2m \times 51.2m \times 64m$ with the voxel grid size of $256 \times 256 \times 32$. By default, the model is trained for 25 epochs. We optimise the process, utilizing the AdamW optimizer with an initial learning rate of 1e-4 and a weight decay of 0.01. We also employ a multi-step scheduler to reduce the learning rate. All models are trained on two A100 Nvidia GPUs with 80G memory and batch size 4.

\paragraph{Implementation of Flow Consistency Check.}\

\textbf{1. Variable Definition:}

\textbf{Forward Flow}: ( \(Flow^{t \rightarrow t-1}\) )

\begin{itemize}
\item
  Maps pixels from the current frame ( \(I_t \)) to the previous frame (
  \(I_{t-1}\) ).
\item
  For a pixel (\({x}_t \in I_t \)), the corresponding location in
  (\(I\_{t-1}\)) is:

  \({x}_{t-1} = {x}_t + Flow^{t \rightarrow t-1}({x}_t)\)
\end{itemize}

\textbf{Backward Flow}: (\( Flow^{t-1 \rightarrow t}\) )

\begin{itemize}
\item
  Maps pixels from the previous frame (\( I_{t-1} \)) to the current
  frame ( \(I_t\) ).
\item
  For a pixel ( \({x}_{t-1} \in I\_{t-1} \)), the corresponding location
  in ( \(I_t\) ) is:

  \({x}_t' = {x}\_{t-1} + Flow^{t-1 \rightarrow t}({x}\_{t-1})\)
\end{itemize}

\textbf{2. Consistency Check}

The \textbf{forward-backward consistency check} verifies whether a pixel
mapping is valid by ensuring round-trip correspondence.

\textbf{Round-trip Mapping}

A pixel in ( \(I_t \)) is mapped to ( \(I_{t-1} \)) using forward flow,
and then mapped back using backward flow:

\({x}_t'' = {x}_t + Flow^{t \rightarrow t-1}({x}_t) + Flow^{t-1 \rightarrow t}({x}_t + F^{t \rightarrow t-1}({x}_t))\)

Define the \textbf{consistency residual} as:\\
\(\Delta({x}_t) = Flow^{t \rightarrow t-1}({x}_t) + Flow^{t-1 \rightarrow t}({x}_t + F^{t \rightarrow t-1}({x}_t))\)

If the norm ( \(\|\Delta({x}_t)\| \)) is small (below a threshold), the
mapping is considered consistent.

\textbf{3. Occlusion Mask}

Pixels with high inconsistency are typically considered \textbf{occluded
or unreliable}.

\textbf{Occlusion Mask (\( M({x}) \)):}

\[M({x}) = 
\begin{cases}
1 & \text{if } \|\Delta({x})\| > \tau \quad \text{(occluded)} \\\\
0 & \text{otherwise} \quad \text{(non-occluded)}
\end{cases}\]

Where (\( \tau \)) is a predefined threshold.

\paragraph{Implementation of FGTA and OGVR module.}
Algorithms~\ref{alg1} and~\ref{alg2} describe the implementation details of the two key components proposed in this work: Flow-Guided Temporal Aggregation (FGTA) and Occlusion-Guided Voxel Refinement (OGVR).

Algorithm~\ref{alg1} outlines the inference procedure for FGTA. Given the current and historical image frames $\{I_{t-i}\}_{i=0}^{N}$ and their corresponding features $\{F_{t-i}\}_{i=0}^{N}$, the algorithm first estimates forward and backward optical flows between the current frame $I_t$ and each historical frame $I_{t-i}$.
Each historical feature map $F_{t-i}$ is warped to the reference frame using flow-guided warping.
Cosine similarity between the warped features and the current frame feature $F_{t}$ is computed to assign adaptive weights, which emphasize temporally consistent and visually similar regions. Simultaneously, an occlusion mask is constructed through a bidirectional flow consistency check to identify unreliable regions. 
The weighted historical features are aggregated into 
$F_{agg}$, and a Neighborhood Cross-Attention (NCA) operation is applied to further refine $F_t$, focusing only on non-occluded regions. This process enhances temporal consistency and robustness to motion and occlusion artifacts.

Algorithm~\ref{alg2} describes the inference procedure for the OGVR module. The goal is to refine the voxel features in 3D space by integrating information from the aggregated feature volume $V_{agg}$, the current frame's voxel features $V_{t}$, and the occlusion mask volume $V_{mask}$. Non-occluded regions are updated using features from $V_{agg}$, while occluded regions are filled in using the current frame's voxel features $V_{t}$, which are more reliable in such areas. A per-voxel weight map is maintained to track the number of valid sources contributing to each voxel. The final voxel feature representation $V_{fine}$is obtained by normalizing the fused features with the accumulated weights, ensuring smooth transitions at the boundaries between occluded and non-occluded regions. Importantly, this refinement process is lightweight and introduces no additional parameter overhead.

Together, these modules enable FlowScene to effectively align temporal information and reason across occlusions in both 2D and 3D spaces, leading to improved semantic and geometric scene understanding in dynamic environments.
\begin{algorithm}
\caption{Inference algorithm of flow-guided temporal alignment}
\label{alg1}
\begin{algorithmic}[1]
\State \textbf{Inputs:} Images $\{I_{t-i}\}_{i=0}^{N}$, Features $\{F_{t-i}\}_{i=0}^{N}$
\State $M= Zeros$ \Comment{init occlusion mask}
\For{$i = 1$ to $N$}
    \State $Flow^{t \rightarrow t-i}, Flow^{t-i \rightarrow t} \gets \mathcal{F}(I_t, I_{t-i})$ \Comment{compute dual optical flow}
    \State $F_{warp}^{t-i\rightarrow t} \gets \mathcal{W}arp(F_{t-i}, Flow^{t \to t-i})$ \Comment{flow-guided warp}
    \State $w_{t-i\rightarrow t} \gets \operatorname{similarity}(F_{warp}^{t-i\rightarrow t},F_t)$ \Comment{compute similarity weight}
    \State ${F}_{agg}[i] \gets w_{t-i\rightarrow t}\cdot F^{t-i\rightarrow t}_{warp}$ \Comment{get the features corresponding to the weights}
    \State $M \gets \mathcal{CC}(Flow^{t \rightarrow t-i}, Flow^{t-i \rightarrow t})\cup M$ \Comment{compute occlusion mask}
\EndFor
\State $F_{agg} \gets \text{SUM}(F_{agg})$
\State $F_t \gets \mathcal{NCA}(F_t,(1-M)\cdot F_{warp})$
\State \textbf{Outputs:} $F_{agg}, F_t, M$
\end{algorithmic}
\end{algorithm}
\begin{algorithm}
\caption{Inference algorithm of occlusion-guided voxel refinement}
\label{alg2}
\begin{algorithmic}[1]
\State \textbf{Inputs:} Aggregated voxel feature $V_{agg}$, 
        current voxel feature $V_t$, 
        occlusion mask $V_{mask}$
\State Initialize $\text{weight} \gets \mathbf{0}$ with shape as $V_t$
\State Initialize $V_{fine} \gets \mathbf{0}$ with shape as $V_t$

\Statex \Comment{fuse non-occluded regions}

\State ${V_{fine}} \gets V_{fine} + V_{agg} \cdot (1 - V_{mask})$

\State $\text{weight} \gets \text{weight} + (1 - V_{mask})$

\Statex \Comment{fuse occluded regions}
\State $V_{fine} \gets V_{fine} + V_{t}$
\State $\text{weight} \gets \text{weight} + \mathbf{1}$

\Statex \Comment{normalize result}
\State $\text{weight} \gets \max(\text{weight}, 1 \times 10^{-6})$
\State $V_{fine} \gets V_{fine} / \text{weight}$

\State \textbf{Outputs:} Refined voxel feature $V_{fine}$
\end{algorithmic}
\end{algorithm}


\begin{table}[h]
  \centering\small
        \caption{Reproduce SOTA method using different image encoder.}
  \begin{tabular}{l|c|ccc}
    \toprule
    {\textbf{Method}}& \textbf{Backbone}&{\textbf{mIoU}(\%)}&{\textbf{Params}(M)} \\
    \midrule
    
    \multirow{2}{*}{MonoScene~\cite{cao2022monoscene}}& EfficientNetB7&12.96&132.4\\
    &RepViT&12.59&91.0\\
    \hline
    \multirow{2}{*}{VoxFormer~\cite{li2023voxformer}}& ResNet50&13.35&57.9\\
    &RepViT&13.62&54.7\\
    \hline
    \multirow{2}{*}{BRGScene~\cite{li2023stereoscene}}& EfficientNetB7&15.43&161.4\\
    &RepViT&16.13&120.0\\
    \hline
    \multirow{2}{*}{HTCL~\cite{li2024htcl}}& EfficientNetB7&17.13&181.4\\
    &RepViT&16.86&140.0\\
    \hline
    \multirow{2}{*}{VLScene~\cite{wang2025vlscene}}& EfficientNetB7&17.44&88.8\\
    &RepViT&\underline{17.83}&\textbf{47.4}\\
    \hline
    \multirow{3}{*}{Ours}& EfficientNetB7&17.63&93.8\\
    &ResNet50&16.98&55.6\\
    &RepViT&\textbf{18.13}&\underline{52.4}\\
    \bottomrule
  \end{tabular}

  \label{tab:flow}
\end{table}

\section{More Results}
\label{results}
\subsection{Reproduce SOTA Method using Different Image Encoder}
\label{Reproduce}
To ensure a fair and consistent comparison across methods, we re-implemented several state-of-the-art Semantic Scene Completion models using a unified experimental setup. Specifically, we replaced the original image encoders used in existing methods with a lightweight and efficient backbone—RepViT~\cite{wang2023repvit}—while keeping all other architectural and training settings unchanged.

As shown in Table~\ref{tab:flow}, the results demonstrate that RepViT significantly reduces the number of parameters across all models, often without degrading performance—and in some cases, even improving it. This confirms the generalizability and efficiency of RepViT as an image encoder for SSC tasks.


\subsection{Standard Errors and Standard Deviations Results}
We conducted a statistical analysis on the SemanticKITTI
validation set and report the \textbf{weighted mean IoU (W-mIoU)},
\textbf{weighted standard deviation (W-SD)}, and \textbf{weighted
standard error (W-SE)} across semantic categories. As shown, our method not only achieves the \textbf{highest mIoU
and W-mIoU}, but also demonstrates \textbf{W-SD} and \textbf{W-SE}
compared to other strong baselines. This indicates that our performance
improvements are {statistically meaningful and stable across
classes}.
\begin{table*}[h]
  \centering
  \caption{Standard Errors and Standard Deviations Results}
  \label{tab:dyna}
  \resizebox{0.75\textwidth}{!}{
  \begin{tabular}{l|cccc}
    \toprule
    Method & mIoU($\uparrow)$ & W-mIoU($\uparrow$) & W-SD($\downarrow$) & W-SE($\downarrow$) \\
    \midrule
    \textbf{Ours} & \textbf{17.70} & \textbf{33.11} & \textbf{13.80} &
    \textbf{6.50} \\
    HTCL (ECCV\textquotesingle2024) & 17.08 & 32.64 & 14.17 & 6.67 \\
    CGFormer (NIPS\textquotesingle2024) & 16.63 & 31.91 & 14.43 & 6.80 \\
    BRGScene (IJCAI\textquotesingle2024) & 15.35 & 30.22 & 14.01 & 6.60 \\
    TPVFormer (ICCV\textquotesingle2023) & 12.32 & 24.44 & 14.34 & 6.75 \\
    MonoScene (CVPR\textquotesingle2022) & 11.08 & 22.58 & 14.38 & 6.77\\
    \bottomrule
  \end{tabular}}
\end{table*}
\subsection{More Quantitative Results}
\label{quantitative}
To provide a more thorough comparison, we provide additional quantitative results of semantic scene completion on the SemanticKITTI validation set in Table~\ref{tab:val}. The results further demonstrate the effectiveness of our approach in enhancing 3D scene perception performance.
Compared with the previous state-of-the-art methods, FlowScene is superior to other HTCL~\cite{li2024htcl} in semantic scene understanding, with a 1.00\% increase in mIoU. In addition, compared with Symphonize~\cite{jiang2024symphonize}, huge improvements are made in both occupancy and semantics. IoU and mIoU enhancement are of great significance for practical applications. It proves that we are not simply reducing a certain metric to achieve semantic scene completion.
\begin{table*}[h]
  \setlength{\tabcolsep}{2pt}
  \renewcommand\arraystretch{1.2}
    \caption{Quantitative results on the SemanticKITTI validation set. The best and the second best results are in \textbf{bold} and \underline{underlined}, respectively.}
  \resizebox{\textwidth}{!}{
  \begin{tabular}{l|l|c|r|rrrrrrrrrrrrrrrrrrr|r}
    \toprule
    \textbf{Methods} &\textbf{Published}&\textbf{Inputs}&\textbf{IoU}   
    & \rotatebox{90}{\vcenteredbox{\colorbox[RGB]{255,0,255}{\textcolor[RGB]{255,0,255}{\rule{1px}{1px}}}} \textbf{road} (15.30$\%$)} 
    & \rotatebox{90}{\vcenteredbox{\colorbox[RGB]{75,0,75}{\textcolor[RGB]{75,0,75}{\rule{1px}{1px}}}} \textbf{sidewalk} (11.13$\%$)} 
    & \rotatebox{90}{\vcenteredbox{\colorbox[RGB]{255,150,255}{\textcolor[RGB]{255,150,255}{\rule{1px}{1px}}}} \textbf{parking} (1.12$\%$)} 
    & \rotatebox{90}{\vcenteredbox{\colorbox[RGB]{175,0,75}{\textcolor[RGB]{175,0,75}{\rule{1px}{1px}}}} \textbf{other-grnd} (0.56$\%$)} 
    &  \rotatebox{90}{\vcenteredbox{\colorbox[RGB]{255,200,0}{\textcolor[RGB]{255,200,0}{\rule{1px}{1px}}}} \textbf{building} (14.10$\%$)} 
    &  \rotatebox{90}{\vcenteredbox{\colorbox[RGB]{100,150,245}{\textcolor[RGB]{100,150,245}{\rule{1px}{1px}}}} \textbf{car} (3.92$\%$)} 
    & \rotatebox{90}{\vcenteredbox{\colorbox[RGB]{80,30,180}{\textcolor[RGB]{80,30,180}{\rule{1px}{1px}}}} \textbf{truck} (0.16$\%$)} 
    & \rotatebox{90}{\vcenteredbox{\colorbox[RGB]{100,230,245}{\textcolor[RGB]{100,230,245}{\rule{1px}{1px}}}} \textbf{bicycle} (0.03$\%$)}  
    & \rotatebox{90}{\vcenteredbox{\colorbox[RGB]{30,60,150}{\textcolor[RGB]{30,60,150}{\rule{1px}{1px}}}} \textbf{motocycle} (0.03$\%$)} 
    & \rotatebox{90}{\vcenteredbox{\colorbox[RGB]{0,0,255}{\textcolor[RGB]{0,0,255}{\rule{1px}{1px}}}} \textbf{other-vehicle} (0.20$\%$)} 
    & \rotatebox{90}{\vcenteredbox{\colorbox[RGB]{0,175,0}{\textcolor[RGB]{0,175,0}{\rule{1px}{1px}}}} \textbf{vegetation} (39.3$\%$)} 
    & \rotatebox{90}{\vcenteredbox{\colorbox[RGB]{135,60,0}{\textcolor[RGB]{135,60,0}{\rule{1px}{1px}}}}  \textbf{trunk} (0.51$\%$)} 
    & \rotatebox{90}{\vcenteredbox{\colorbox[RGB]{150,240,80}{\textcolor[RGB]{150,240,80}{\rule{1px}{1px}}}} \textbf{terrain} (9.17$\%$)} 
    & \rotatebox{90}{\vcenteredbox{\colorbox[RGB]{255,30,30}{\textcolor[RGB]{255,30,30}{\rule{1px}{1px}}}} \textbf{person} (0.07$\%$)} 
    & \rotatebox{90}{\vcenteredbox{\colorbox[RGB]{255,40,200}{\textcolor[RGB]{255,40,200}{\rule{1px}{1px}}}} \textbf{bicylist} (0.07$\%$)} 
    & \rotatebox{90}{\vcenteredbox{\colorbox[RGB]{150,30,90}{\textcolor[RGB]{150,30,90}{\rule{1px}{1px}}}}  \textbf{motorcyclist} (0.05$\%$)} 
    &  \rotatebox{90}{\vcenteredbox{\colorbox[RGB]{255,120,50}{\textcolor[RGB]{255,120,50}{\rule{1px}{1px}}}} \textbf{fence} (3.90$\%$)} 
    & \rotatebox{90}{\vcenteredbox{\colorbox[RGB]{255,240,150}{\textcolor[RGB]{255,240,150}{\rule{1px}{1px}}}} \textbf{pole} (0.29$\%$)} 
    & \rotatebox{90}{\vcenteredbox{\colorbox[RGB]{255,0,0}{\textcolor[RGB]{255,0,0}{\rule{1px}{1px}}}} \textbf{traf.-sign} (0.08$\%$)}& \textbf{mIoU}  \\
    
    \midrule
    MonoScene \cite{cao2022monoscene}&CVPR'2022 &S& 36.86&56.52&26.72&14.27&0.46&14.09&23.26&6.98&0.61&0.45&1.48&17.89&2.81&29.64&1.86&1.20&0.00&5.84&4.14&2.25&11.08 \\
    
    TPVFormer \cite{huang2023tri}& CVPR'2023&S& 35.61&56.50& 25.87& 20.60& 0.85& 13.88& 23.81& 8.08& 0.36& 0.05 &4.35 &16.92& 2.26& 30.38& 0.51& 0.89& 0.00& 5.94& 3.14& 1.52&11.36 \\
    
    OccFormer\cite{zhang2023occformer}& ICCV'2023 &S& 36.50  & 58.85& 26.88& 19.61& 0.31& 14.40& 25.09& \underline{25.53}& 0.81& 1.19& 8.52& 19.63& 3.93& 32.62& 2.78& \underline{2.82}& 0.00& 5.61& 4.26& 2.86 & 13.46\\
    Symphonize~\cite{jiang2024symphonize}&CVPR'2024&S&41.92&56.37&27.58&15.28&0.95&21.64&28.68&20.44&\underline{2.54}&2.82&\underline{13.89}&25.72&6.60&30.87&\underline{3.52}&2.24&0.00&8.40&9.57&5.76&14.89\\
        
    VoxFormer-T\cite{li2023voxformer}& CVPR'2023&T& 44.15 &53.57&26.52&19.69&0.42&19.54&26.54&7.26&1.28&0.56&7.81&26.10&6.10&33.06&1.93&1.97&0.00&7.31&9.15&4.94 & 13.35\\
    H2GFormer~\cite{wang2024h2gformer} & AAAI'2024&T&44.69&57.00&29.37&21.74&0.34&20.51&28.21&6.80&0.95&0.91&9.32&\textbf{27.44}&7.80&36.26&1.15&0.10&0.00&7.98&9.88&5.81&14.29 \\
    HASSC~\cite{wang2024HASSC}&CVPR'2024&T&44.58&55.30&29.60&\textbf{25.90}&\textbf{11.30}&23.10&23.00&2.90&1.90&1.50&4.90&24.80&\textbf{9.80}&26.50&1.40&\textbf{3.00}&0.00&\textbf{14.30}&7.00&\textbf{7.10}&14.74\\
    HTCL~\cite{li2024htcl}&ECCV'2024&T&\textbf{45.51}&\underline{63.70}&\textbf{32.48}&\underline{23.27}&0.14&\underline{24.13}&\textbf{34.30}&20.72&\textbf{3.99}&2.80&11.99&\underline{26.96}&\underline{8.79}&\textbf{37.73}&2.56&2.70&0.00&11.22&\underline{11.49}&6.95&\underline{17.13}\\
    \midrule
    \rowcolor{gray!20}\textbf{Ours}&  &T& {\underline{45.01}} &{\textbf{63.72}}&{\underline{32.10}}& 22.20&\underline{1.31}& {\textbf{25.63}}& {\underline{33.33}}&\textbf{33.47}&2.36& {\textbf{5.09}}&\textbf{16.99}& {{26.35}}& {8.68}& {\underline{36.73}}&\textbf{3.79}& {{1.92}}& {\textbf{0.00}}& {\underline{12.05}}& {\textbf{11.65}}& {\underline{7.05}}&  {\textbf{18.13}}\\
    \bottomrule
  \end{tabular}}

  \label{tab:val}
\end{table*}
\subsection{Failure Case}
\begin{figure*}[t]
\centering
  \includegraphics[width=\textwidth]{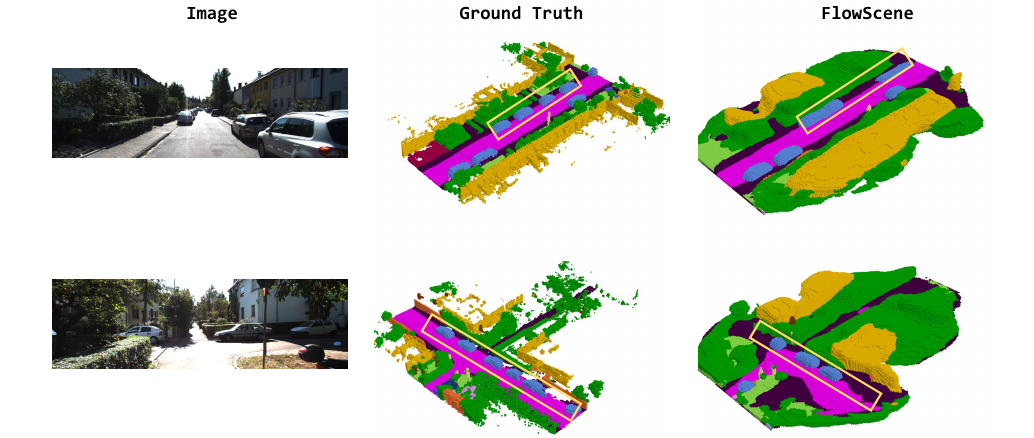}
  \caption{Failure cases.}
  \label{fig:sup3}
\end{figure*}

We provide two failure cases in Figure~\ref{fig:sup3}.

\subsection{More Visualizations Results}
\label{visualizations}
\begin{figure*}[t]
\centering
  \includegraphics[width=\textwidth]{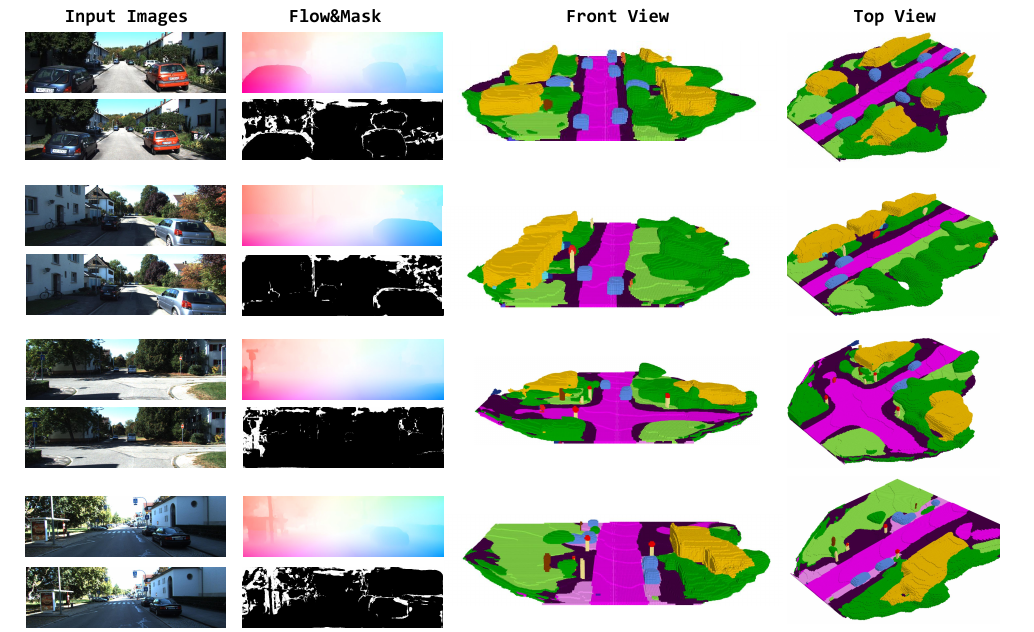}
  \caption{Qualitative results on the SemanticKITTI validation set.}
  \label{fig:sup1}
\end{figure*}

\begin{figure*}[t]
\centering
  \includegraphics[width=\textwidth]{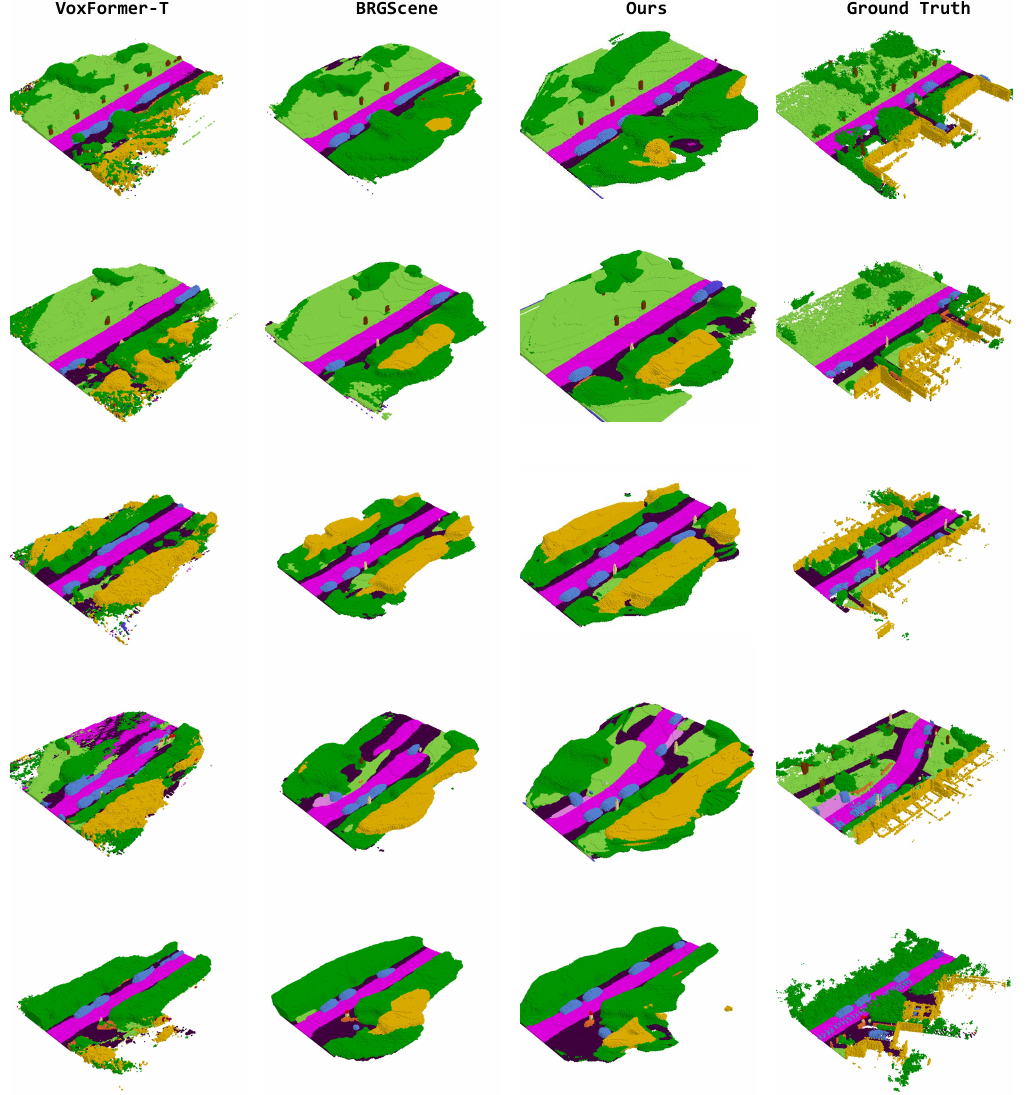}
  \caption{Qualitative results on the SemanticKITTI validation set.}
  \label{fig:sup2}
\end{figure*}

We show visualization examples on the Semantickitti validation set, as shown in Figure~\ref{fig:sup1}. From left to right are the input image, the corresponding optical flow and occlusion mask, the front view SSC, and the top view SSC. Due to the motion information brought by the optical flow, the location information of the scene objects is more accurate and the layout is more reasonable.
We report the performance of more visual comparison results on the SemanticKITTI validation set in Figure~\ref{fig:sup2}. We compare with VoxFormer~\cite{li2023voxformer} and BRGScene~\cite{li2023stereoscene}. In general, our method performs more fine-grained segmentation of the scene and maintains clear segmentation boundaries. For example, in the segmentation completion result of cars, we predict clear separation of each car. In contrast, other methods show continuous semantic errors for occluded cars. In addition, our flow can effectively deal with the problem of mutual occlusion between different objects. Finally, we provide a video in the appendix to show the performance more intuitively.

\section{Discussions}
\label{sec:limit}
\subsection{Limitations}
Flowscene shows strong performance on the benchmark with an improved number of parameters. This is beneficial for deploying real-world autonomous driving applications. But the inference time of the model needs to be improved. While optical flow is effective, it depends on pretrained flow model, potentially limiting performance in degraded visual conditions.
\subsection{Future Works}
Semantic scene completion in multi-camera settings is also worth attention, which is our future work. Meanwhile, the legal challenges of autonomous driving as well as privacy and data security risks are still topics of debate. Finally, the robustness of semantic scene completion is also an issue worth exploring.
\subsection{Broader Impacts}
FlowScene enhances 3D geometry perception by aligning temporal features through optical flow information, thereby improving the ability of temporal semantic scene completion. This work has a non-obvious negative social impact.

\clearpage



\newpage
\section*{NeurIPS Paper Checklist}
\begin{enumerate}

\item {\bf Claims}
    \item[] Question: Do the main claims made in the abstract and introduction accurately reflect the paper's contributions and scope?
    \item[] Answer: \answerYes{} 
    \item[] Justification: We have described our contributions and scope explicitly in both the abstract and introduction.
    \item[] Guidelines:
    \begin{itemize}
        \item The answer NA means that the abstract and introduction do not include the claims made in the paper.
        \item The abstract and/or introduction should clearly state the claims made, including the contributions made in the paper and important assumptions and limitations. A No or NA answer to this question will not be perceived well by the reviewers. 
        \item The claims made should match theoretical and experimental results, and reflect how much the results can be expected to generalize to other settings. 
        \item It is fine to include aspirational goals as motivation as long as it is clear that these goals are not attained by the paper. 
    \end{itemize}

\item {\bf Limitations}
    \item[] Question: Does the paper discuss the limitations of the work performed by the authors?
    \item[] Answer: \answerYes{} 
    \item[] Justification: We describe our limitations in the section "Discussions" in our appendix text.
    \item[] Guidelines:
    \begin{itemize}
        \item The answer NA means that the paper has no limitation while the answer No means that the paper has limitations, but those are not discussed in the paper. 
        \item The authors are encouraged to create a separate "Limitations" section in their paper.
        \item The paper should point out any strong assumptions and how robust the results are to violations of these assumptions (e.g., independence assumptions, noiseless settings, model well-specification, asymptotic approximations only holding locally). The authors should reflect on how these assumptions might be violated in practice and what the implications would be.
        \item The authors should reflect on the scope of the claims made, e.g., if the approach was only tested on a few datasets or with a few runs. In general, empirical results often depend on implicit assumptions, which should be articulated.
        \item The authors should reflect on the factors that influence the performance of the approach. For example, a facial recognition algorithm may perform poorly when image resolution is low or images are taken in low lighting. Or a speech-to-text system might not be used reliably to provide closed captions for online lectures because it fails to handle technical jargon.
        \item The authors should discuss the computational efficiency of the proposed algorithms and how they scale with dataset size.
        \item If applicable, the authors should discuss possible limitations of their approach to address problems of privacy and fairness.
        \item While the authors might fear that complete honesty about limitations might be used by reviewers as grounds for rejection, a worse outcome might be that reviewers discover limitations that aren't acknowledged in the paper. The authors should use their best judgment and recognize that individual actions in favor of transparency play an important role in developing norms that preserve the integrity of the community. Reviewers will be specifically instructed to not penalize honesty concerning limitations.
    \end{itemize}

\item {\bf Theory assumptions and proofs}
    \item[] Question: For each theoretical result, does the paper provide the full set of assumptions and a complete (and correct) proof?
    \item[] Answer: \answerNA{} 
    \item[] Justification: This paper does not include theoretical results.
    \item[] Guidelines:
    \begin{itemize}
        \item The answer NA means that the paper does not include theoretical results. 
        \item All the theorems, formulas, and proofs in the paper should be numbered and cross-referenced.
        \item All assumptions should be clearly stated or referenced in the statement of any theorems.
        \item The proofs can either appear in the main paper or the supplemental material, but if they appear in the supplemental material, the authors are encouraged to provide a short proof sketch to provide intuition. 
        \item Inversely, any informal proof provided in the core of the paper should be complemented by formal proofs provided in appendix or supplemental material.
        \item Theorems and Lemmas that the proof relies upon should be properly referenced. 
    \end{itemize}

    \item {\bf Experimental result reproducibility}
    \item[] Question: Does the paper fully disclose all the information needed to reproduce the main experimental results of the paper to the extent that it affects the main claims and/or conclusions of the paper (regardless of whether the code and data are provided or not)?
    \item[] Answer: \answerYes{} 
    \item[] Justification: We have provided implementation details in our "Experiments" section. Moreover, the source code will be released upon acceptance.
    \item[] Guidelines:
    \begin{itemize}
        \item The answer NA means that the paper does not include experiments.
        \item If the paper includes experiments, a No answer to this question will not be perceived well by the reviewers: Making the paper reproducible is important, regardless of whether the code and data are provided or not.
        \item If the contribution is a dataset and/or model, the authors should describe the steps taken to make their results reproducible or verifiable. 
        \item Depending on the contribution, reproducibility can be accomplished in various ways. For example, if the contribution is a novel architecture, describing the architecture fully might suffice, or if the contribution is a specific model and empirical evaluation, it may be necessary to either make it possible for others to replicate the model with the same dataset, or provide access to the model. In general. releasing code and data is often one good way to accomplish this, but reproducibility can also be provided via detailed instructions for how to replicate the results, access to a hosted model (e.g., in the case of a large language model), releasing of a model checkpoint, or other means that are appropriate to the research performed.
        \item While NeurIPS does not require releasing code, the conference does require all submissions to provide some reasonable avenue for reproducibility, which may depend on the nature of the contribution. For example
        \begin{enumerate}
            \item If the contribution is primarily a new algorithm, the paper should make it clear how to reproduce that algorithm.
            \item If the contribution is primarily a new model architecture, the paper should describe the architecture clearly and fully.
            \item If the contribution is a new model (e.g., a large language model), then there should either be a way to access this model for reproducing the results or a way to reproduce the model (e.g., with an open-source dataset or instructions for how to construct the dataset).
            \item We recognize that reproducibility may be tricky in some cases, in which case authors are welcome to describe the particular way they provide for reproducibility. In the case of closed-source models, it may be that access to the model is limited in some way (e.g., to registered users), but it should be possible for other researchers to have some path to reproducing or verifying the results.
        \end{enumerate}
    \end{itemize}

\item {\bf Open access to data and code}
    \item[] Question: Does the paper provide open access to the data and code, with sufficient instructions to faithfully reproduce the main experimental results, as described in supplemental material?
    \item[] Answer: \answerYes{} 
    \item[] Justification: The source code will be released upon acceptance. One can easily reproduce our results after preparing the SemanticKITTI and SSCBench-KITTI-360 datasets as required.
    \item[] Guidelines:
    \begin{itemize}
        \item The answer NA means that paper does not include experiments requiring code.
        \item Please see the NeurIPS code and data submission guidelines (\url{https://nips.cc/public/guides/CodeSubmissionPolicy}) for more details.
        \item While we encourage the release of code and data, we understand that this might not be possible, so “No” is an acceptable answer. Papers cannot be rejected simply for not including code, unless this is central to the contribution (e.g., for a new open-source benchmark).
        \item The instructions should contain the exact command and environment needed to run to reproduce the results. See the NeurIPS code and data submission guidelines (\url{https://nips.cc/public/guides/CodeSubmissionPolicy}) for more details.
        \item The authors should provide instructions on data access and preparation, including how to access the raw data, preprocessed data, intermediate data, and generated data, etc.
        \item The authors should provide scripts to reproduce all experimental results for the new proposed method and baselines. If only a subset of experiments are reproducible, they should state which ones are omitted from the script and why.
        \item At submission time, to preserve anonymity, the authors should release anonymized versions (if applicable).
        \item Providing as much information as possible in supplemental material (appended to the paper) is recommended, but including URLs to data and code is permitted.
    \end{itemize}

\item {\bf Experimental setting/details}
    \item[] Question: Does the paper specify all the training and test details (e.g., data splits, hyperparameters, how they were chosen, type of optimizer, etc.) necessary to understand the results?
    \item[] Answer: \answerYes{} 
    \item[] Justification: Yes, our training/test details are detailedly presented in our "Experimental Setup" section.
    \item[] Guidelines:
    \begin{itemize}
        \item The answer NA means that the paper does not include experiments.
        \item The experimental setting should be presented in the core of the paper to a level of detail that is necessary to appreciate the results and make sense of them.
        \item The full details can be provided either with the code, in appendix, or as supplemental material.
    \end{itemize}

\item {\bf Experiment statistical significance}
    \item[] Question: Does the paper report error bars suitably and correctly defined or other appropriate information about the statistical significance of the experiments?
    \item[] Answer: \answerNo{} 
    \item[] Justification: Error bars are not reported because it would be too computationally expensive in this paper.
    \item[] Guidelines:
    \begin{itemize}
        \item The answer NA means that the paper does not include experiments.
        \item The authors should answer "Yes" if the results are accompanied by error bars, confidence intervals, or statistical significance tests, at least for the experiments that support the main claims of the paper.
        \item The factors of variability that the error bars are capturing should be clearly stated (for example, train/test split, initialization, random drawing of some parameter, or overall run with given experimental conditions).
        \item The method for calculating the error bars should be explained (closed form formula, call to a library function, bootstrap, etc.)
        \item The assumptions made should be given (e.g., Normally distributed errors).
        \item It should be clear whether the error bar is the standard deviation or the standard error of the mean.
        \item It is OK to report 1-sigma error bars, but one should state it. The authors should preferably report a 2-sigma error bar than state that they have a 96\% CI, if the hypothesis of Normality of errors is not verified.
        \item For asymmetric distributions, the authors should be careful not to show in tables or figures symmetric error bars that would yield results that are out of range (e.g. negative error rates).
        \item If error bars are reported in tables or plots, The authors should explain in the text how they were calculated and reference the corresponding figures or tables in the text.
    \end{itemize}

\item {\bf Experiments compute resources}
    \item[] Question: For each experiment, does the paper provide sufficient information on the computer resources (type of compute workers, memory, time of execution) needed to reproduce the experiments?
    \item[] Answer: \answerYes{} 
    \item[] Justification: The computation resources are detailed in the "Experimental Setup" section.
    \item[] Guidelines: 
    \begin{itemize}
        \item The answer NA means that the paper does not include experiments.
        \item The paper should indicate the type of compute workers CPU or GPU, internal cluster, or cloud provider, including relevant memory and storage.
        \item The paper should provide the amount of compute required for each of the individual experimental runs as well as estimate the total compute. 
        \item The paper should disclose whether the full research project required more compute than the experiments reported in the paper (e.g., preliminary or failed experiments that didn't make it into the paper). 
    \end{itemize}
    
\item {\bf Code of ethics}
    \item[] Question: Does the research conducted in the paper conform, in every respect, with the NeurIPS Code of Ethics \url{https://neurips.cc/public/EthicsGuidelines}?
    \item[] Answer: \answerYes{} 
    \item[] Justification: Our work adheres to the NeurIPS Code of Ethics.
    \item[] Guidelines: 
    \begin{itemize}
        \item The answer NA means that the authors have not reviewed the NeurIPS Code of Ethics.
        \item If the authors answer No, they should explain the special circumstances that require a deviation from the Code of Ethics.
        \item The authors should make sure to preserve anonymity (e.g., if there is a special consideration due to laws or regulations in their jurisdiction).
    \end{itemize}

\item {\bf Broader impacts}
    \item[] Question: Does the paper discuss both potential positive societal impacts and negative societal impacts of the work performed?
    \item[] Answer: \answerYes{} 
    \item[] Justification: We discuss the societal impacts of our work in the "Discussions" section in the appendix
    \item[] Guidelines:
    \begin{itemize}
        \item The answer NA means that there is no societal impact of the work performed.
        \item If the authors answer NA or No, they should explain why their work has no societal impact or why the paper does not address societal impact.
        \item Examples of negative societal impacts include potential malicious or unintended uses (e.g., disinformation, generating fake profiles, surveillance), fairness considerations (e.g., deployment of technologies that could make decisions that unfairly impact specific groups), privacy considerations, and security considerations.
        \item The conference expects that many papers will be foundational research and not tied to particular applications, let alone deployments. However, if there is a direct path to any negative applications, the authors should point it out. For example, it is legitimate to point out that an improvement in the quality of generative models could be used to generate deepfakes for disinformation. On the other hand, it is not needed to point out that a generic algorithm for optimizing neural networks could enable people to train models that generate Deepfakes faster.
        \item The authors should consider possible harms that could arise when the technology is being used as intended and functioning correctly, harms that could arise when the technology is being used as intended but gives incorrect results, and harms following from (intentional or unintentional) misuse of the technology.
        \item If there are negative societal impacts, the authors could also discuss possible mitigation strategies (e.g., gated release of models, providing defenses in addition to attacks, mechanisms for monitoring misuse, mechanisms to monitor how a system learns from feedback over time, improving the efficiency and accessibility of ML).
    \end{itemize}
    
\item {\bf Safeguards}
    \item[] Question: Does the paper describe safeguards that have been put in place for responsible release of data or models that have a high risk for misuse (e.g., pretrained language models, image generators, or scraped datasets)?
    \item[] Answer: \answerNA{} 
    \item[] Justification: The involved data/models does not pose a high risk for misuse.
    \item[] Guidelines:
    \begin{itemize}
        \item The answer NA means that the paper poses no such risks.
        \item Released models that have a high risk for misuse or dual-use should be released with necessary safeguards to allow for controlled use of the model, for example by requiring that users adhere to usage guidelines or restrictions to access the model or implementing safety filters. 
        \item Datasets that have been scraped from the Internet could pose safety risks. The authors should describe how they avoided releasing unsafe images.
        \item We recognize that providing effective safeguards is challenging, and many papers do not require this, but we encourage authors to take this into account and make a best faith effort.
    \end{itemize}

\item {\bf Licenses for existing assets}
    \item[] Question: Are the creators or original owners of assets (e.g., code, data, models), used in the paper, properly credited and are the license and terms of use explicitly mentioned and properly respected?
    \item[] Answer: \answerYes{} 
    \item[] Justification: In our paper, we primarily engage with public datasets, we have cited them properly and set the license in the website of the OpenReview.
    \item[] Guidelines:
    \begin{itemize}
        \item The answer NA means that the paper does not use existing assets.
        \item The authors should cite the original paper that produced the code package or dataset.
        \item The authors should state which version of the asset is used and, if possible, include a URL.
        \item The name of the license (e.g., CC-BY 4.0) should be included for each asset.
        \item For scraped data from a particular source (e.g., website), the copyright and terms of service of that source should be provided.
        \item If assets are released, the license, copyright information, and terms of use in the package should be provided. For popular datasets, \url{paperswithcode.com/datasets} has curated licenses for some datasets. Their licensing guide can help determine the license of a dataset.
        \item For existing datasets that are re-packaged, both the original license and the license of the derived asset (if it has changed) should be provided.
        \item If this information is not available online, the authors are encouraged to reach out to the asset's creators.
    \end{itemize}

\item {\bf New assets}
    \item[] Question: Are new assets introduced in the paper well documented and is the documentation provided alongside the assets?
    \item[] Answer: \answerNA{} 
    \item[] Justification: This paper does not release new assets.
    \item[] Guidelines:
    \begin{itemize}
        \item The answer NA means that the paper does not release new assets.
        \item Researchers should communicate the details of the dataset/code/model as part of their submissions via structured templates. This includes details about training, license, limitations, etc. 
        \item The paper should discuss whether and how consent was obtained from people whose asset is used.
        \item At submission time, remember to anonymize your assets (if applicable). You can either create an anonymized URL or include an anonymized zip file.
    \end{itemize}

\item {\bf Crowdsourcing and research with human subjects}
    \item[] Question: For crowdsourcing experiments and research with human subjects, does the paper include the full text of instructions given to participants and screenshots, if applicable, as well as details about compensation (if any)? 
    \item[] Answer: \answerNA{} 
    \item[] Justification: The paper does not involve crowdsourcing nor research with human subjects.
    \item[] Guidelines:
    \begin{itemize}
        \item The answer NA means that the paper does not involve crowdsourcing nor research with human subjects.
        \item Including this information in the supplemental material is fine, but if the main contribution of the paper involves human subjects, then as much detail as possible should be included in the main paper. 
        \item According to the NeurIPS Code of Ethics, workers involved in data collection, curation, or other labor should be paid at least the minimum wage in the country of the data collector. 
    \end{itemize}

\item {\bf Institutional review board (IRB) approvals or equivalent for research with human subjects}
    \item[] Question: Does the paper describe potential risks incurred by study participants, whether such risks were disclosed to the subjects, and whether Institutional Review Board (IRB) approvals (or an equivalent approval/review based on the requirements of your country or institution) were obtained?
    \item[] Answer: \answerNA{} 
    \item[] Justification: The paper does not involve crowdsourcing nor research with human subjects.
    \item[] Guidelines:
    \begin{itemize}
        \item The answer NA means that the paper does not involve crowdsourcing nor research with human subjects.
        \item Depending on the country in which research is conducted, IRB approval (or equivalent) may be required for any human subjects research. If you obtained IRB approval, you should clearly state this in the paper. 
        \item We recognize that the procedures for this may vary significantly between institutions and locations, and we expect authors to adhere to the NeurIPS Code of Ethics and the guidelines for their institution. 
        \item For initial submissions, do not include any information that would break anonymity (if applicable), such as the institution conducting the review.
    \end{itemize}

\item {\bf Declaration of LLM usage}
    \item[] Question: Does the paper describe the usage of LLMs if it is an important, original, or non-standard component of the core methods in this research? Note that if the LLM is used only for writing, editing, or formatting purposes and does not impact the core methodology, scientific rigorousness, or originality of the research, declaration is not required.
    \item[] Answer: \answerNA{} 
    \item[] Justification: Our method development in this research does not involve LLMs as any important, original, or non-standard components.
    \item[] Guidelines:
    \begin{itemize}
        \item The answer NA means that the core method development in this research does not involve LLMs as any important, original, or non-standard components.
        \item Please refer to our LLM policy (\url{https://neurips.cc/Conferences/2025/LLM}) for what should or should not be described.
    \end{itemize}

\end{enumerate}

\end{document}